\documentclass[conference]{IEEEtran}
\IEEEoverridecommandlockouts
\usepackage{cite}
\usepackage{amsmath,amssymb,amsfonts}
\usepackage{algorithmic}
\usepackage{graphicx}
\usepackage{textcomp}
\usepackage{xcolor}
\usepackage{subfigure}

\def\BibTeX{{\rm B\kern-.05em{\sc i\kern-.025em b}\kern-.08em
    T\kern-.1667em\lower.7ex\hbox{E}\kern-.125emX}}
\begin{document}

\title{Detecting the Anomalies in LiDAR Pointcloud\\
}

\author{\IEEEauthorblockN{Chiyu Zhang, Ji Han, Yao Zou, Kexin Dong, Yujia Li, Junchun Ding, Xiaoling Han}
\IEEEauthorblockA{
\texttt{\{firstname.lastname\}@tusimple.ai} \\
\textit{TuSimple}\\
\textit{San Diego, CA, 92122}
}
}

\maketitle

\begin{abstract}
    LiDAR sensors play an important role in the perception stack of modern autonomous driving systems.
Adverse weather conditions such as rain, fog and dust, as well as some (occasional) LiDAR hardware fault may cause the LiDAR to produce pointcloud with abnormal patterns such as scattered noise points and uncommon intensity values. 
In this paper, we propose a novel approach to detect whether a LiDAR is generating anomalous pointcloud by analyzing the pointcloud characteristics. 
Specifically, we develop a pointcloud quality metric based on the LiDAR points' spatial and intensity distribution to characterize the noise level of the pointcloud, which relies on pure mathematical analysis and does not require any labeling or training as learning-based methods do. Therefore, the method is scalable and can be quickly deployed either online to improve the autonomy safety by monitoring anomalies in the LiDAR data or offline to perform in-depth study of the LiDAR behavior over large amount of data.
The proposed approach is studied with extensive real public road data collected by LiDARs with different scanning mechanisms and laser spectrums, and is proven to be able to effectively handle various known and unknown sources of pointcloud anomaly.

\end{abstract}


\section{Introduction}
\label{sec:intro}
LiDAR (Light Detection and Ranging) sensors have caught growing attention of the automotive and autonomous driving industry thanks to their capability of continuously generating high-definition and accurately-ranged image (pointcloud) of the surroundings, regardless of the ambient illuminance conditions~\cite{royo2019overview,li2020lidar}.
As is pointed out in~\cite{li2020lidar}, one particular challenge of using LiDARs for perception in autonomous driving is the performance degradation in adverse weather conditions such as rain, fog, dust, etc., where the LiDAR's laser signal may be scattered and/or attenuated, leading to reduced laser power and signal-noise ratio (SNR) and thus may cause the pointcloud to contain random noise points and lower intensity readings~\cite{bijelic2018benchmark}.
Not only the adverse environmental conditions can cause the issues above, sometimes defected LiDAR hardware components or unknown random factors may also lead to anomalous pointcloud output. 
For example, a LiDAR with defected electromagnetic shielding may output extremely noisy pointcloud when strong signal interference sources such as cellular towers are nearby. 
The goal of this paper is to propose a method to characterize the aforementioned LiDAR pointcloud anomalies, which can benefit the autonomous driving system (ADS) safety as well as the ADS development cycle.
In terms of increasing the level of automation and the ADS safety, a higher level ADS (level 3+) needs to detect whether the system is within its operation domain and behave correspondingly, according to the Society of Automotive Engineers (SAE)~\cite{sae2021standard}. The ADS operation domain is typically bounded by environmental conditions and system component health, and it is essential that the ADS sensors such as LiDARs are able to determine their status and data quality. 
As for the application in the ADS development, the data frames with anomalous LiDAR pointcloud are typically associated with edge cases and long-tail scenarios, which require extra attention yet have relatively low rate of occurrence in the vast amount of data generated by the autonomous driving fleet. Having those cases picked out effectively and efficiently helps to save the time and effort required for ADS development.

While researches on general LiDAR pointcloud anomalies are limited, the topic of LiDAR performance under adverse weather conditions have been studied extensively~\cite{al2004fog,ryde2009performance,rasshofer2011influences,hasirlioglu2016test,filgueira2017quantifying,kutila2018automotive,goodin2019predicting,montalban2021quantitative}.
Many of the studies focus on the performance degradation of the LiDAR in rain/fog and have developed various quantification methods for aspects such as signal attenuation, visibility range, point density and target reflectance. 
Some recent studies develop statistical-based learning methods to classify whether a LiDAR is working in adverse weather based on performance degradation metrics~\cite{heinzler2019weather,zhang2021lidar}. 
These methods are typically verified through simulation or testing in controlled environment which may not well resemble the realistic road conditions. For example, many controlled environments to emulate rains such as the one presented in~\cite{heinzler2019weather} consists of several static test targets (vehicles, pedestrians, etc.). Such environment cannot produce water splashes generated from rolling wheels of other vehicles on the road, which is typically seen and picked up by the LiDARs in realistic operations. 
In addition, it should be noted that many of the commonly studied LiDAR performance degradation aspects do not always lead to safety-critical component or system failure.
For example, a LiDAR typically have a reduced visibility range in rain which only reduces the perception system's capability and does not necessarily disable all the perception functions; on the other hand, even if the LiDAR is operating with its full capability in a sunny day, it may generate a large amount of false positive points due to hardware failure which is likely to be recognized as objects by the perception system and cause the vehicle to perform a hard-brake.
In~\cite{ruff2019deep} the authors developed a deep-learning based approach to classify and detect LiDAR pointcloud anomalies. However, there are two major drawbacks to apply the deep-learning based approaches in practical R\&D and implementation. First, it requires a large amount of annotated LiDAR data frames to train the software, moreover, the data collection, annotation and training pipeline must be repeated for different LiDAR properties, such as spinning vs solid state, 905nm vs 1550nm, or even a change to the mounting locations, thus lengthens the R\&D cycle; and second, the real-time computational cost is high and may not be desirable given the limited onboard computational cost. 

In this paper, we propose a novel quality metric to quantitatively characterize the general noise-related anomalies in LiDAR pointcloud. 
To capture the spatially-scattered nature of LiDAR noise points, we adopt the idea of spatial autocorrelation~\cite{moran1950notes}, which is widely used in statistical studies, to quantify how `dispersed' the points are in a frame of LiDAR pointcloud.
A factor related to the intensity of the pointcloud is also included in the quality metric to better separate the cases where the LiDAR is in heavy rain or dense fog.
The main contribution of the paper is twofold:
\begin{itemize}
    \item First, we developed a general quality metric that is able to capture noise-related anomalies in LiDAR pointcloud regardless of the cause of the anomaly. It is particularly useful in identifying new pointcloud issues with unknown causes or very little prior experience during both early-stage system validation or large-scaled operation. 
    \item Second, the proposed approach does not require a priori data collection, labeling and training and thus can reduce the time and resource consumption for practical implementation.
\end{itemize}
The proposed quality metric is verified with over 10,000 miles of public road data collected by LiDARs with various laser spectrums, scanning mechanisms and mounting locations. The results show that the proposed method is able to identify the pointcloud affected not only by adverse weather conditions, but also by uncommon noise sources such as signal interference, road dust, etc.

The rest of the paper is organized as follows. 
We first present the formulation and implementation of the proposed LiDAR pointcloud quality metric in Section~\ref{sec:method}.
Section~\ref{sec:result} demonstrates the verification of the proposed method, followed by conclusions in Section~\ref{sec:conclusion}.

\section{Pointcloud Quality Metric}
\label{sec:method}
In this section, we first showcase some typical scenarios and characteristics of anomalous LiDAR pointcloud, based on which we formulate the pointcloud quality metric. An implementation method utilizing LiDAR image grid and GPU (graphic processing unit) acceleration is also presented.

\subsection{Anomalous LiDAR Pointcloud}
\label{subsec:lidar-pc}

\begin{figure}[htb]
    \centering
    \subfigure[rain example 1]{
    \includegraphics[width=0.2\textwidth]{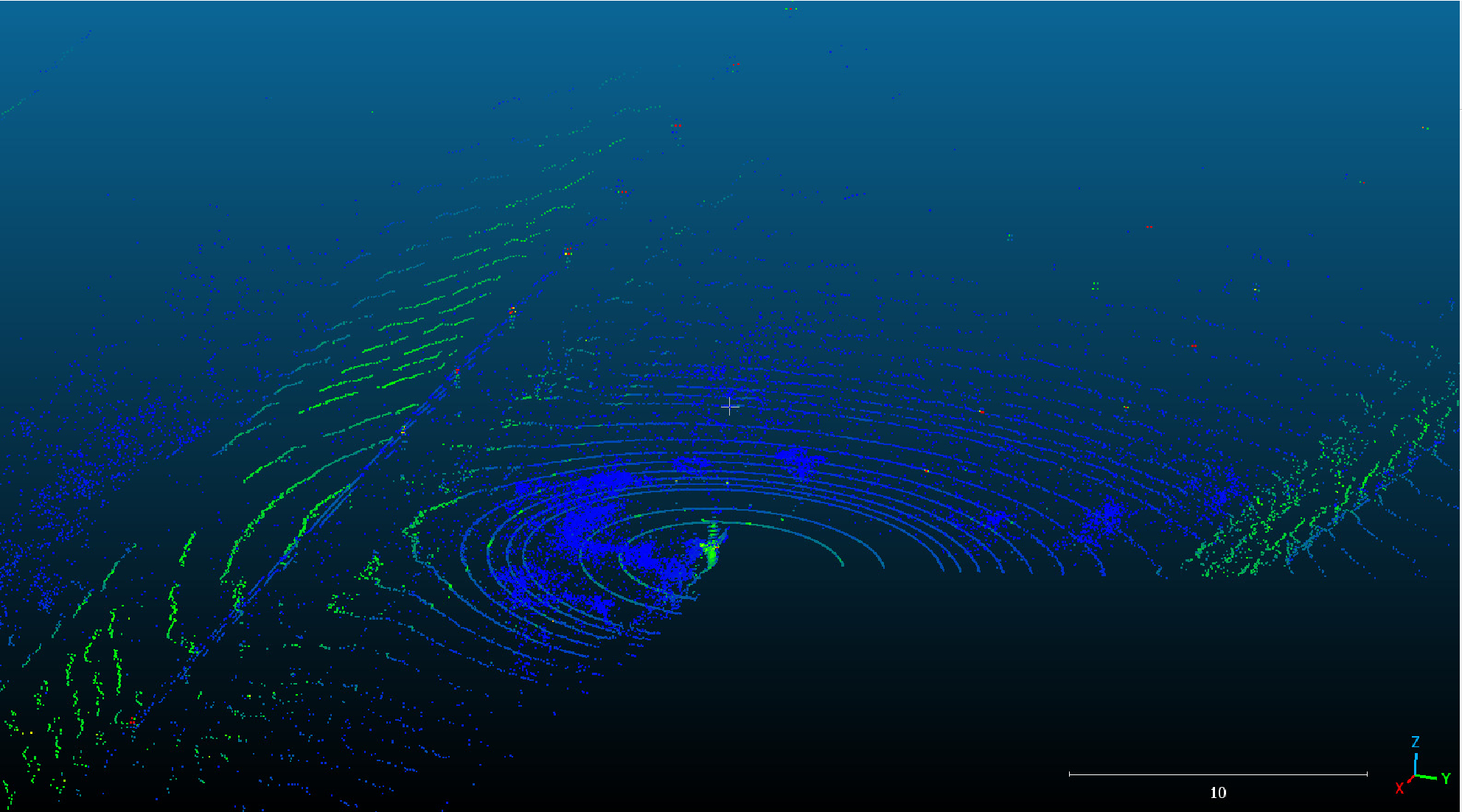}
    \label{fig:pandar-rain}
    }
    \subfigure[rain example 2]{
    \includegraphics[width=0.2\textwidth]{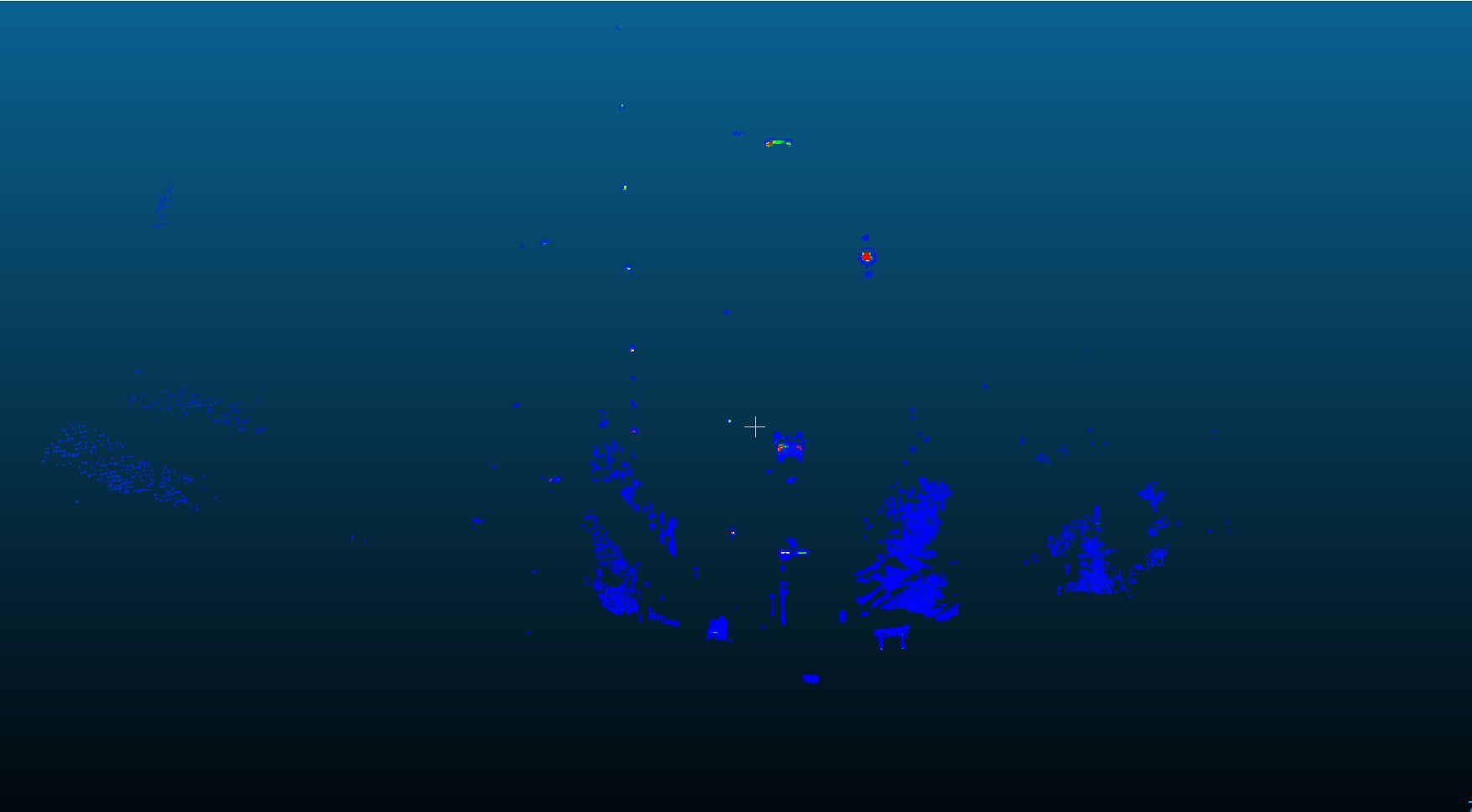}
    \label{fig:inno-rain}
    }
    \subfigure[hardware failure example]{
    \includegraphics[width=0.2\textwidth]{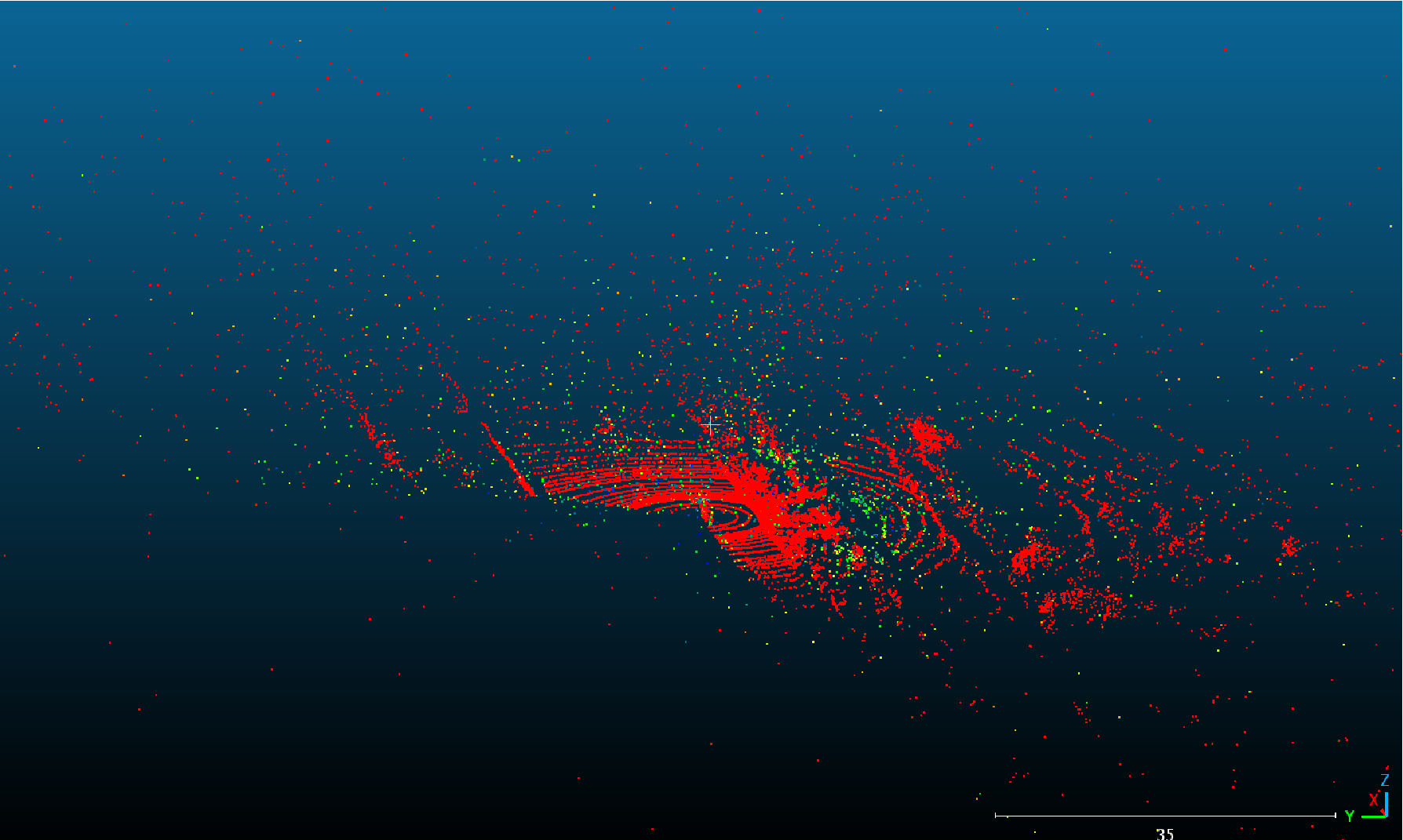}
    \label{fig:pandar-failure}
    }
    \subfigure[interference example]{
    \includegraphics[width=0.2\textwidth]{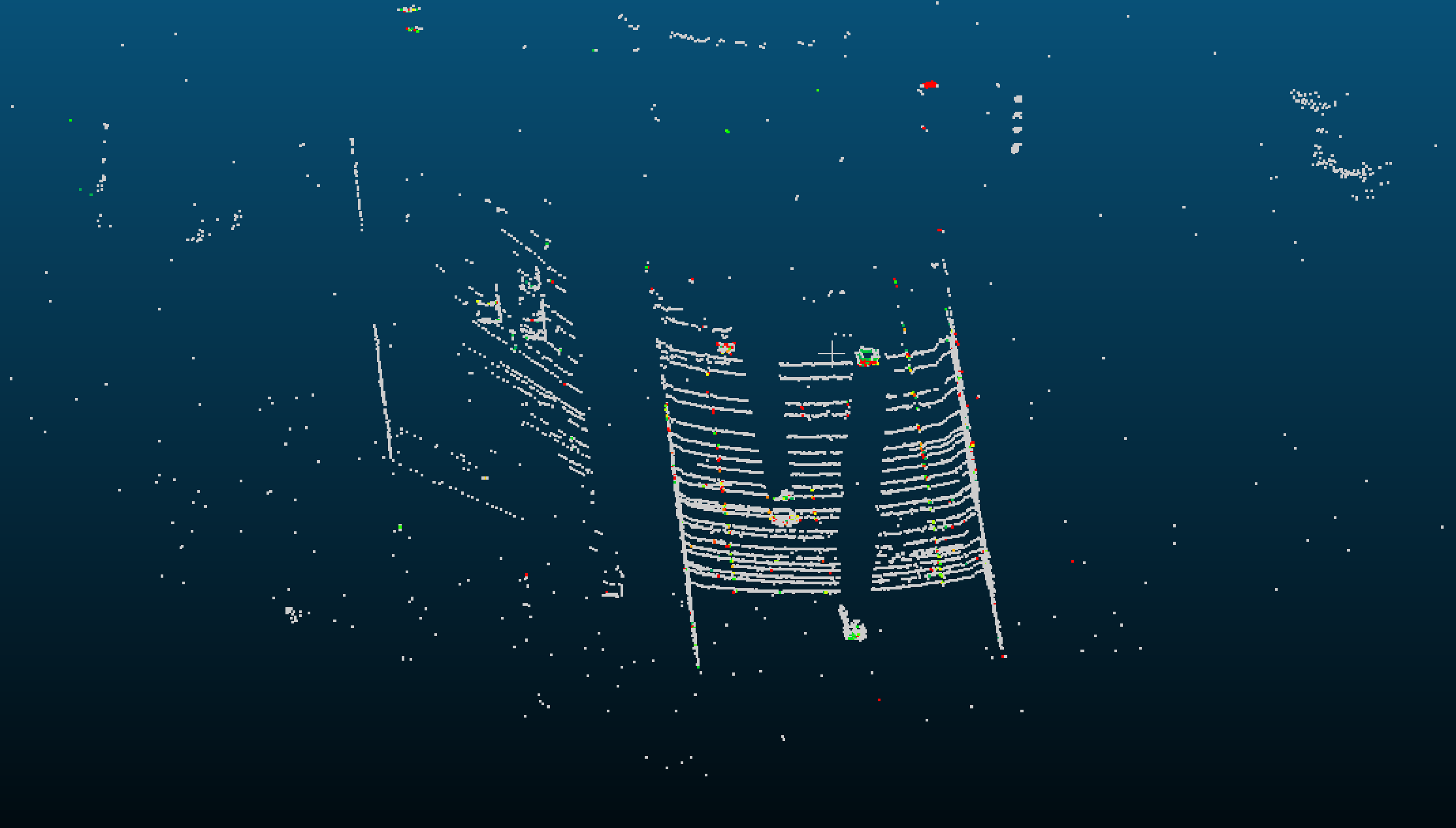}
    \label{fig:inno-failure}
    }
    \caption{Examples of Anomalous LiDAR Pointcloud}
    \label{fig:unknown-s2gmphd}
\end{figure}

LiDAR pointcloud impacted by adverse weathers or hardware component failures may produce anomalous pointcloud with the following typical characteristics:
\begin{itemize}
    \item Randomly and sparsely distributed detections in the 3-dimensional physical space. Signal interference and hardware failure typically affect the LiDAR's signal processing module and generate random and sparse false positives. In adverse weather conditions, this is mainly caused by reflection from water droplets, reflection from scattered laser signals through water/dust, and reduced pointcloud density due to signal attenuation.
    \item Abnormal intensity values. Particularly in rainy and foggy weathers, the intensity values are lower than normal due to signal attenuation. Signal interference and hardware failure may lead to either low or excessively high intensity values.
\end{itemize}
A few examples of typical anomalous LiDAR pointcloud we collected during public road testing are shown in Figure~\ref{fig:unknown-s2gmphd}. All the pointcloud in the figures are colored by the intensity values. Points colored blue indicate low intensity values and those colored red represent high intensity values.
Figure~\ref{fig:pandar-rain} demonstrates one case of LiDAR pointcloud in rain where numerous noise points can be observed at a close range of the LiDAR's field of view (FOV). 
Figure~\ref{fig:inno-rain} shows another case of LiDAR pointcloud in rain. In this case, both the number of points and the intensity values are  significantly reduced due to laser signal getting absorbed by the heavy rain.
The pointcloud in Figure~\ref{fig:pandar-failure} sees much higher intensity values as well as noise points all over the FOV due to an internal component failure inside the LiDAR. 
The LiDAR whose pointcloud is shown Figure~\ref{fig:inno-failure} does not have proper electromagnetic shielding and suffers signal interference when passing a cellular signal tower.

\subsection{Pointcloud quality metric Formulation}
\label{subsec:metric-form}
The proposed pointcloud quality metric consists of two factors to address the two major characteristics of anomalous LiDAR pointcloud shown above. 
The first factor is a spatial measure to quantify how dispersed the LiDAR points are distributed in the 3-dimensional physical space. 
The second factor is an intensity measure to capture the abnormal intensity pattern in the LiDAR pointcloud, particularly the lower-than-normal intensity values in adverse weather conditions such as rain and fog.

\subsubsection{Spatial Measure}
\begin{figure}[htb]
    \centering
    \includegraphics[width=0.25\textwidth]{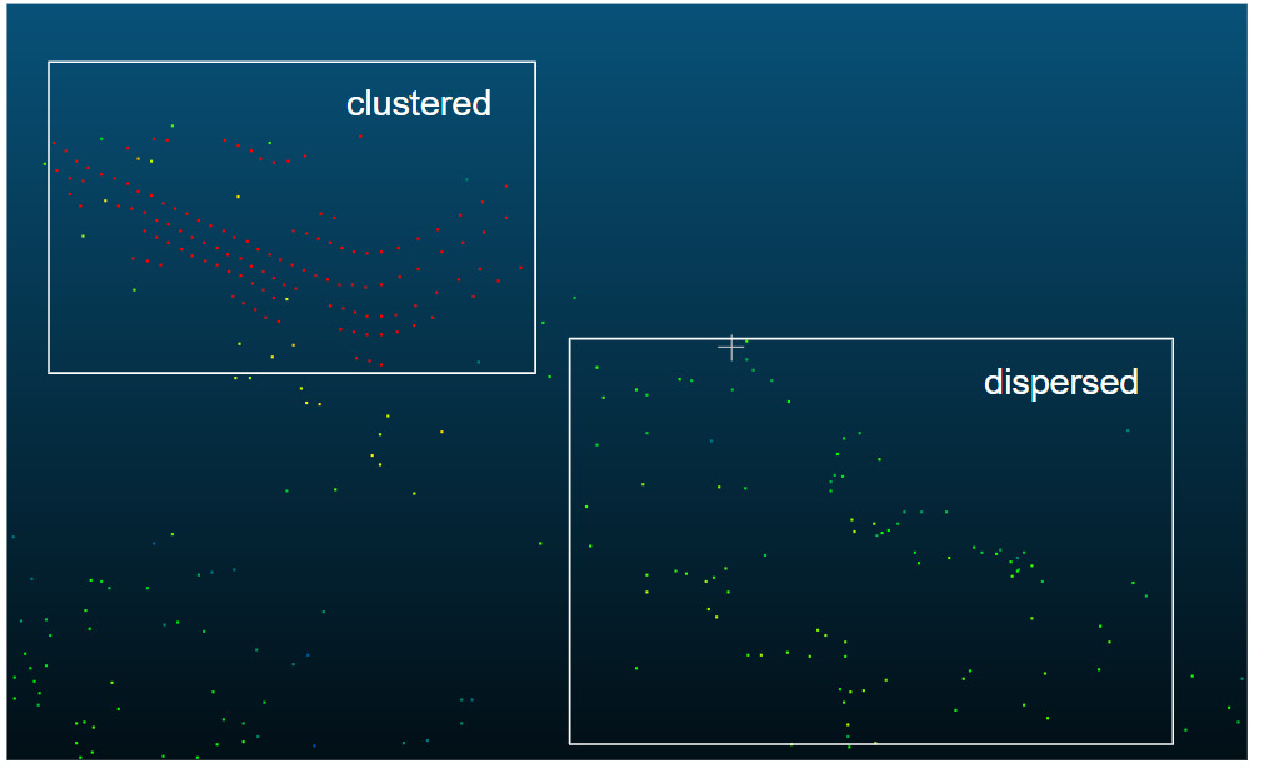}
    \caption{Illustration of Clustered and Dispersed LiDAR Pointcloud}
    \label{fig:pc-example}
\end{figure}

We employ the concept of spatial autocorrelation~\cite{moran1950notes} as a measure of the LiDAR points' level of spatial dispersion. 
In statistics, spatial autocorrelation is used to describe the overall spatial clustering of a group of data by calculating each data point's correlation with other nearby data points. 
A low spatial autocorrelation means that the group of data is dispersed, while a high spatial autocorrelation means that the data group is clustered. 
%
%
The underlying idea of using spatial autocorrelation to characterize the LiDAR pointcloud's spatial dispersion/clustering is that if a segment of LiDAR pointcloud data is generated by lasers detecting an actual object, the distance values in the data segment tend to be clustered since common road objects such as cars and pedestrians typically have large and continuous reflection surfaces. On the other hand, if a LiDAR data segment contains an excessive number of noise points, the distance values in the data segment are more likely dispersed. 
An illustration of the idea is shown in Figure~\ref{fig:pc-example}. The example captures the LiDAR pointcloud of a vehicle driving on wet road surfaces with water splash generated at the rear of the vehicle. The LiDAR points from the vehicle (marked red) are well clustered, while the water splash points behind the vehicle (marked green) are dispersed.

The spatial autocorrelation of a set of LiDAR points is defined as follows. Given a set of LiDAR points:

\begin{equation}
    \mathcal{P}=\{p_i=(r_i,\theta_i,\phi_i,\gamma_i)|i=1,2,...N\}
    \label{eqn:point-set}
\end{equation}

where $r_i,~\theta_i,~\phi_i$ and $\gamma_i$ represents the distance, azimuth, elevation and intensity of the i-th LiDAR point, respectively. Then, the spatial autocorrelation of the distance values is defined as:

\begin{equation}
    I = \begin{cases}
        \hspace{2mm} \dfrac{N}{W}\dfrac{\sum_{i=1}^N\sum_{j=1}^Nw_{ij}(r_i-r)(r_j-r)}{\sum_{i=1}^N(r_i-r)^2} \hspace{5mm} & N>1 \\
        \hspace{2mm} -1 & N=1
        \end{cases}
    \label{eqn:moran}
\end{equation}

$r=\frac{1}{N}\sum_{i=1}^Nr_i$ is the average distance of all distance values in the set of points. $w_{ij}$ is a pre-defined weight value. 
For instance, one may consider the correlation of one data point to all other data points in the set with identical weights by defining $w_{ij}$ as:

\begin{equation}
    w_{ij} = \begin{cases}
        \hspace{2mm} 1 \hspace{5mm} & i\ne j \\
        \hspace{2mm} 0 & i=j
        \end{cases}
    \label{eqn:weight1}
\end{equation}

Alternatively, $w_{ij}$ can also be defined based on the inverse angular distance between points $i$ and $j$ so that the correlation between closer points have higher weight:

\begin{equation}
    w_{ij} = \begin{cases}
        \hspace{2mm} ||(\theta_i,\phi_i),(\theta_j,\phi_j)||^{-2} \hspace{5mm} & i\ne j \\
        \hspace{2mm} 0 & i=j
        \end{cases}
    \label{eqn:weight2}
\end{equation}

$W=\sum_{i=1}^N\sum_{j=1}^Nw_{ij}$ is the sum of all weights. The spatial autocorrelation is valued between $[-1,~1]$, where a value of -1 indicates that the set of points are extremely dispersed in the 3-dimensional physical space and a value of 1 means that the points are well clustered. 
It should be noted that by the definition above, a set with one isolated point, i.e., $N=1$, is considered as dispersed and has a spatial autocorrelation value of -1. 
We believe that (\ref{eqn:moran}) is a reasonable definition for isolated points since an isolated point is most likely to be treated as a noise point in perception algorithms.

\begin{figure}[htb]
    \centering
    \subfigure[case i]{
    \includegraphics[width=0.2\textwidth]{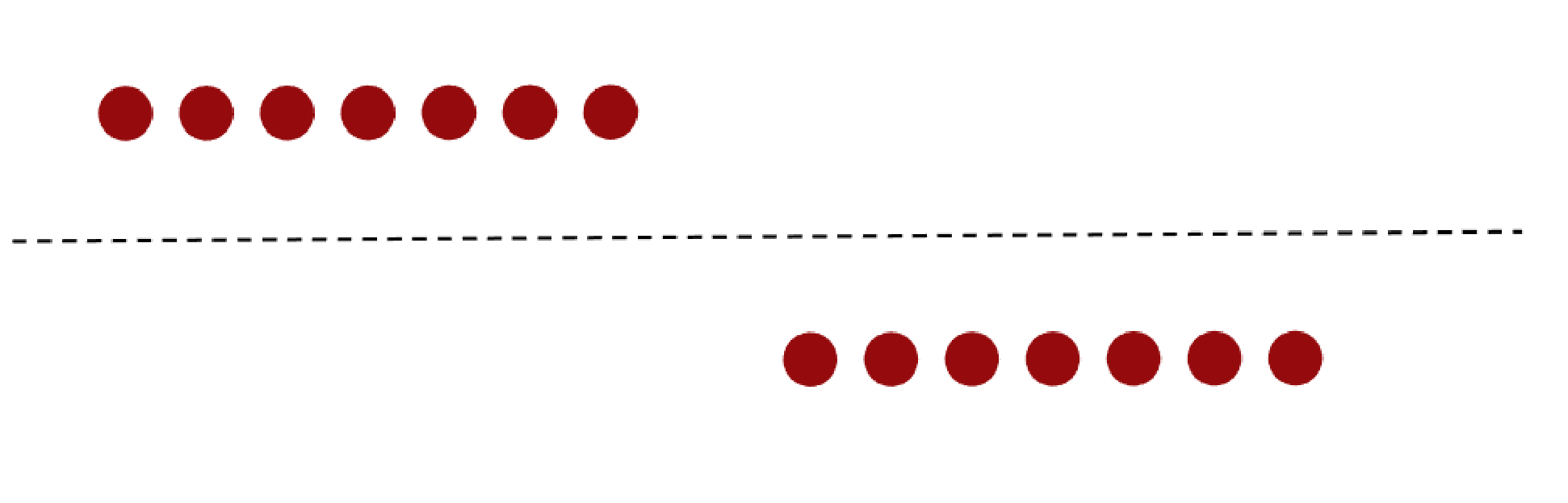}
    \label{fig:autocorr-case1}
    }
    \subfigure[case ii]{
    \includegraphics[width=0.2\textwidth]{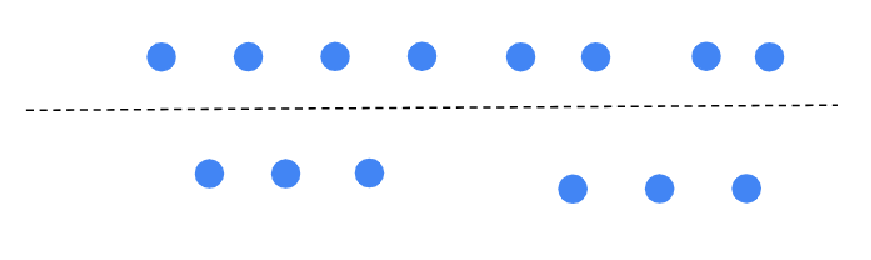}
    \label{fig:autocorr-case2}
    }
    \caption{Examples of Pointcloud Distribution}
    \label{fig:autocorr-cases}
\end{figure}

The main difference between the autocorrelation and statistical variance is that the statistical variance only considers the absolute difference between each individual points to the average, thus, it depicts how the data is distributed in the sample space. 
The spatial autocorrelation, on the other hand, considers the relation between each individual points to other points. Sets of data points that have the same statistical variance may not necessarily have the same spatial autocorrelation.
As shown by the two pointcloud examples in Figure~\ref{fig:autocorr-cases}, where both sets of points shall have the same range variance. However, the spatial autocorrelation of the pointcloud in case ii is negative while that in case i is positive, indicating that the pointcloud in case ii is more dispersed. 
In practice, multiple vehicles/objects in the LiDAR field of view can typically generate a pointcloud distribution similar to case i, and noise/false positives may result in a pointcloud distribution which resembles that in case ii. 
Furthermore, consider the extreme case where only one isolated LiDAR point is present. By definition, the single-point set has a minimum variance of 0. On the other hand, it has the lowest spacial autocorrelation score following the definition~(\ref{eqn:moran}), which aligns with our intention to characterize isolated points as noise points. 
Therefore, spatial autocorrelation is a more suitable measure for our application than the statistical variance.

\subsubsection{Intensity Measure}
\begin{figure}[htb]
    \centering
    \includegraphics[width=0.2\textwidth]{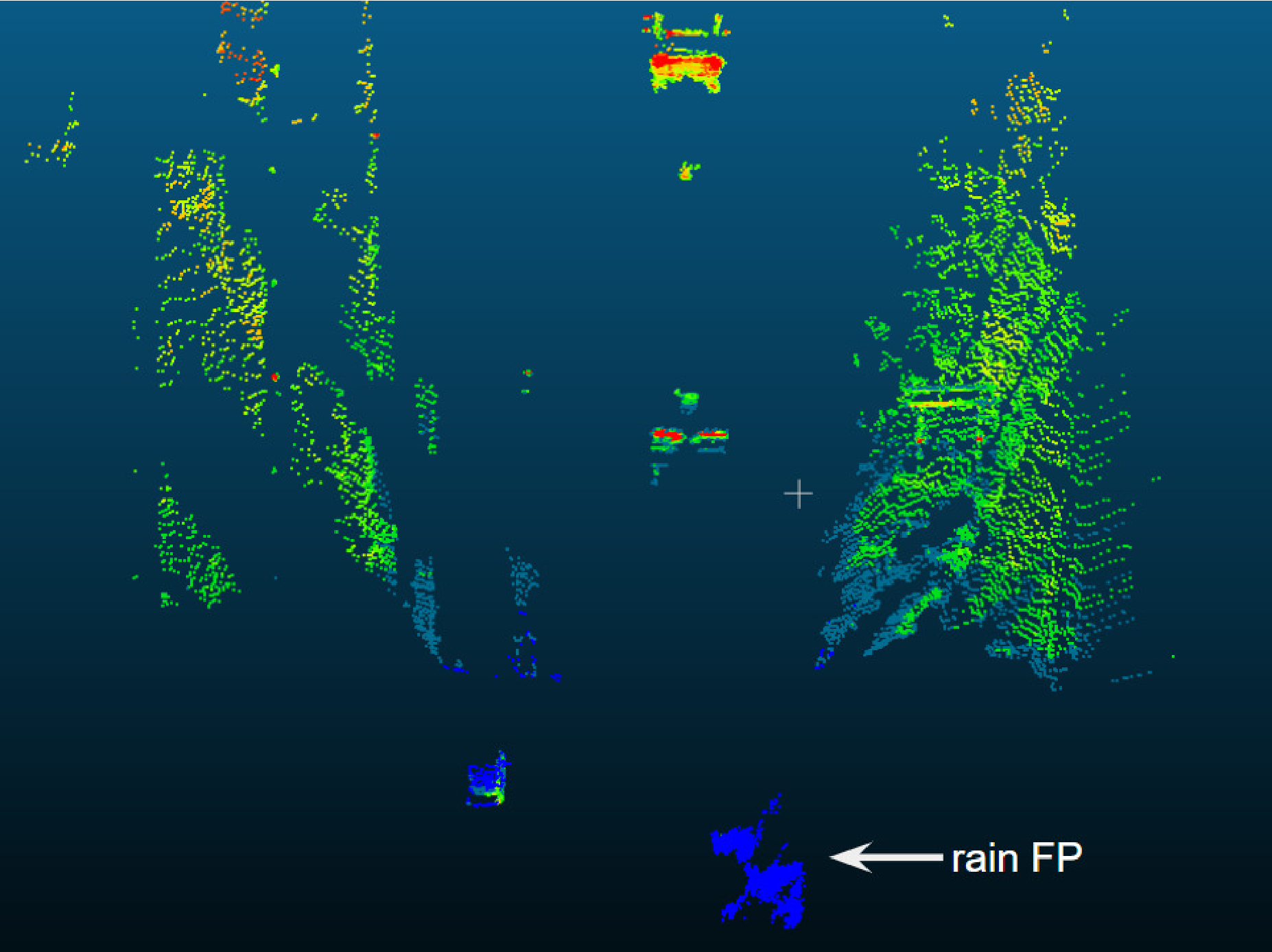}
    \caption{Exemplary LiDAR Noise in Heavy Rain}
    \label{fig:rain-example}
\end{figure}

LiDARs with specific laser wavelengths may generate clustered instead of scattered noise points in heavy rain of dense fog.
Figure~\ref{fig:rain-example} shows an example of such type of noise points. The pointcloud in the figure is captured when the LiDAR encounters heavy rain on the road. In the lower right of the figure there is a sizable cluster of noise points likely generated from reflections of rain droplets, which could be recognized as an object to be avoided by the perception algorithms. 

Since this particular type of LiDAR noise is typically clustered, it can be hard to characterize using spatial autocorrelation alone, as will be shown in the test results in Section~\ref{sec:result}.
However, we have observed that this noise type only occurs when there is a dense layer of laser-absorbing/deflecting matter such as heavy rain, dense fog or intense smog, etc., and the points almost always have extremely low intensity values since they are generated from partial reflection of the laser pulse passing through the matter. 
Therefore, in addition to the spatial autocorrelation, we also take low intensity values into consideration by adding an intensity weight multiplier to the spatial autocorrelation. The intensity weight multiplier can be formulated from any intensity statistical measures such as mean, standard deviation, or any other metrics that can distinguish the abnormally low intensity values. In this paper, we present one formulation of the intensity multiplier based on the average intensity. 

\begin{figure}[htb]
    \centering
    \includegraphics[width=0.3\textwidth]{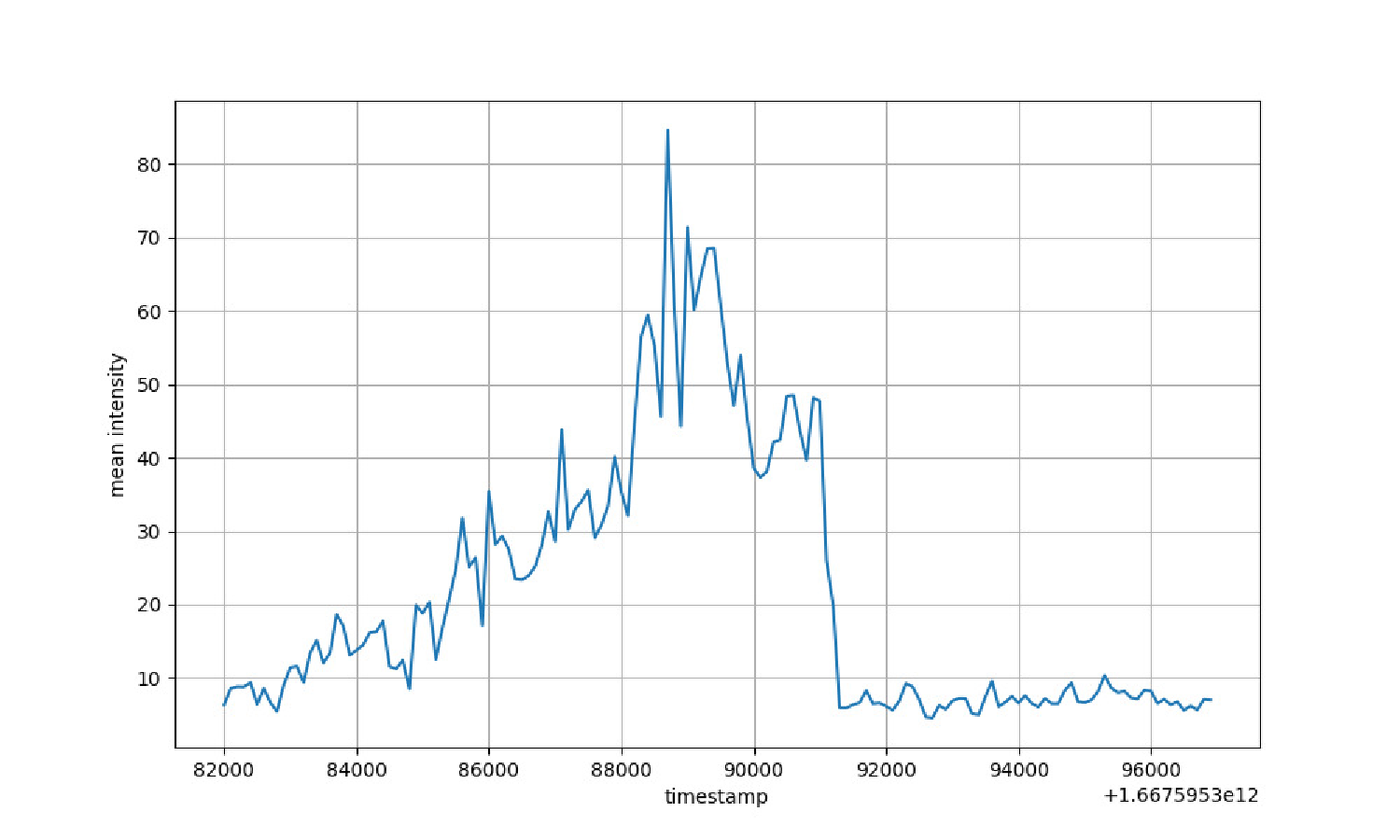}
    \caption{Average Intensity of LiDAR Passing Road Sign}
    \label{fig:intensity-example}
\end{figure}

Let $\gamma_{ref}$ be a reference intensity value which indicates a nominal LiDAR intensity during normal operation (clear weather, no hardware issues). 
The reference is a user-defined value which is typically associated to specific LiDAR models from different manufacturers. The reference value can be obtained through statistical analysis of LiDAR data, since the LiDAR intensity during normal operation is typically consistent with small fluctuations. 
Let $\bar\gamma$ be the average intensity of the set of LiDAR point $\mathcal{P}$. The intensity weight multiplier $K_\gamma$ is formulated as below:

\begin{equation}
    K_\gamma = exp(k\cdot\frac{max(0,\gamma_{ref}-\bar\gamma)}{\gamma_{ref}})
    \label{eqn:intensity-mult}
\end{equation}

where $k$ is a constant scale factor. 
By definition, a low average intensity leads to a high weight multiplier. 
The multiplier value is defined as 1 for high average intensities. While some LiDAR hardware failures may lead to a high average intensity in some cases, as shown in Figure~\ref{fig:pandar-failure}, most of the high average intensity cases are the result of retro-reflective targets, e.g., road signs, occurring at a close range and occupies most of the LiDAR pointcloud. 
Figure~\ref{fig:intensity-example} shows an example of the average intensity of the pointcloud from one test LiDAR passing a road sign. The average intensity ramps up as the road sign gets closer to the vehicle and producing more points. Once the road sign gets out of the LiDAR's FOV, the average intensity quickly drops back to its nominal value. 

These cases with high average intensities are irrelevant to the LiDAR data quality yet are very commonly seen as vehicles can pass road signs from time to time. Therefore, we intentionally disregard the high average intensity in the definition of the multiplier.
Overall, the LiDAR data quality metric is formulated as the multiplication of the intensity weight multiplier and the spatial autocorrelation $K_\gamma\cdot I$. 

\subsection{Implementation}
\label{subsec:metric-implementation}

\subsubsection{LiDAR Image Grid}
It makes practical sense to calculate the spatial autocorrelation of the LiDAR points in a small local area instead of calculating for all LiDAR points across the entire FOV all at once, since typical objects and other physical features do not occupy the entire LiDAR FOV and the LiDAR points are bound to be scattered when looking from a global FOV perspective. 
Furthermore, calculating in a small local area reduces the computational cost as the spatial autocorrelation is of $O(N^2)$ with $N$ being the size of the pointcloud under consideration. 
Therefore, in implementation, we first create a LiDAR image grid and calculate the spatial autocorrelation grid by grid.

\begin{figure}[htb]
    \centering
    \includegraphics[width=0.4\textwidth]{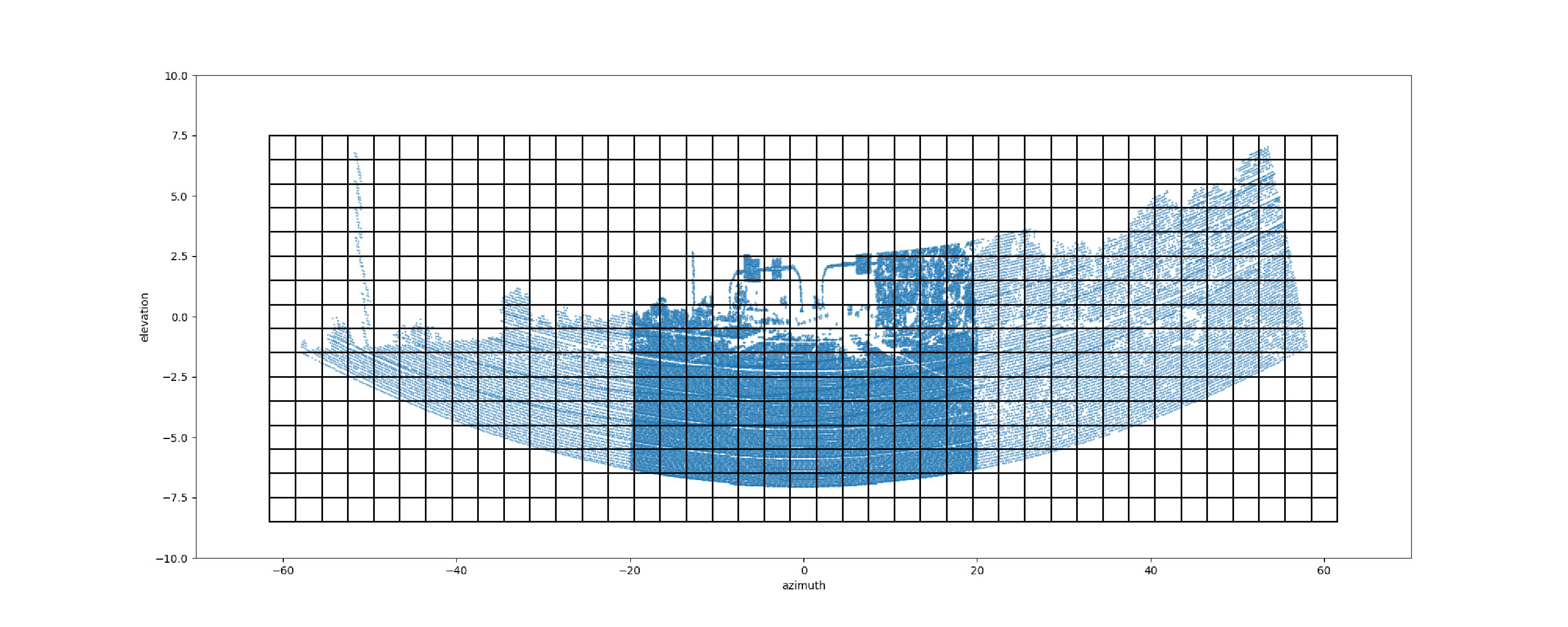}
    \caption{Example of LiDAR Image Grid}
    \label{fig:grid-example}
\end{figure}

For each LiDAR data frame, we project all the LiDAR points onto an azimuth-elevation image, with each point containing its range and intensity information.
The image is then divided into grids in both azimuth and elevation directions. An example of such image grid is shown In Figure~\ref{fig:grid-example}.
Then, for each grid cell, we calculate the weighted spatial autocorrelation of all the distance values of the points in that cell following the definition~(\ref{eqn:moran}) and~(\ref{eqn:intensity-mult}). 
The overall quality metric score of the LiDAR data frame is then the sum of the weighted spatial autocorrelation over all grid cells averaged by the number of grid cells:

\begin{equation}
    s = \dfrac{1}{VH}\displaystyle\sum_i^V\displaystyle\sum_j^H K_{\gamma,ij}\cdot I_{ij}
    \label{eqn:grid-moran}
\end{equation}

where $i$ and $j$ denotes the indices of the grid cells, and the $V$ and $H$ denotes the number of grid cells in the elevation and azimuth directions, respectively. 

\subsubsection{GPU Acceleration}
By definition, the time complexity to calculate the spatial autocorrelation is of $O(N^2)$, where $N$ is the number of point a LiDAR produces in one frame. Therefore, the time cost of calculating the weighted spatial autocorrelation can be too high to meet the real-time constraint since modern automotive LiDARs can generate up to 100,000 points in a single frame.
Applying the implementation based on the LiDAR pointcloud image grid shown above, the computation can be done in parallel for each grid cell since the spatial autocorrelation of each grid cell is independent to other grid cells.
As GPUs become a more and more viable resource on automotives~\cite{ditty2023systems}, in this section, we propose a GPU-accelerated parallel computation implementation of the weighted spatial autocorrelation.

\begin{figure}[htb]
    \centering
    \includegraphics[width=0.3\textwidth]{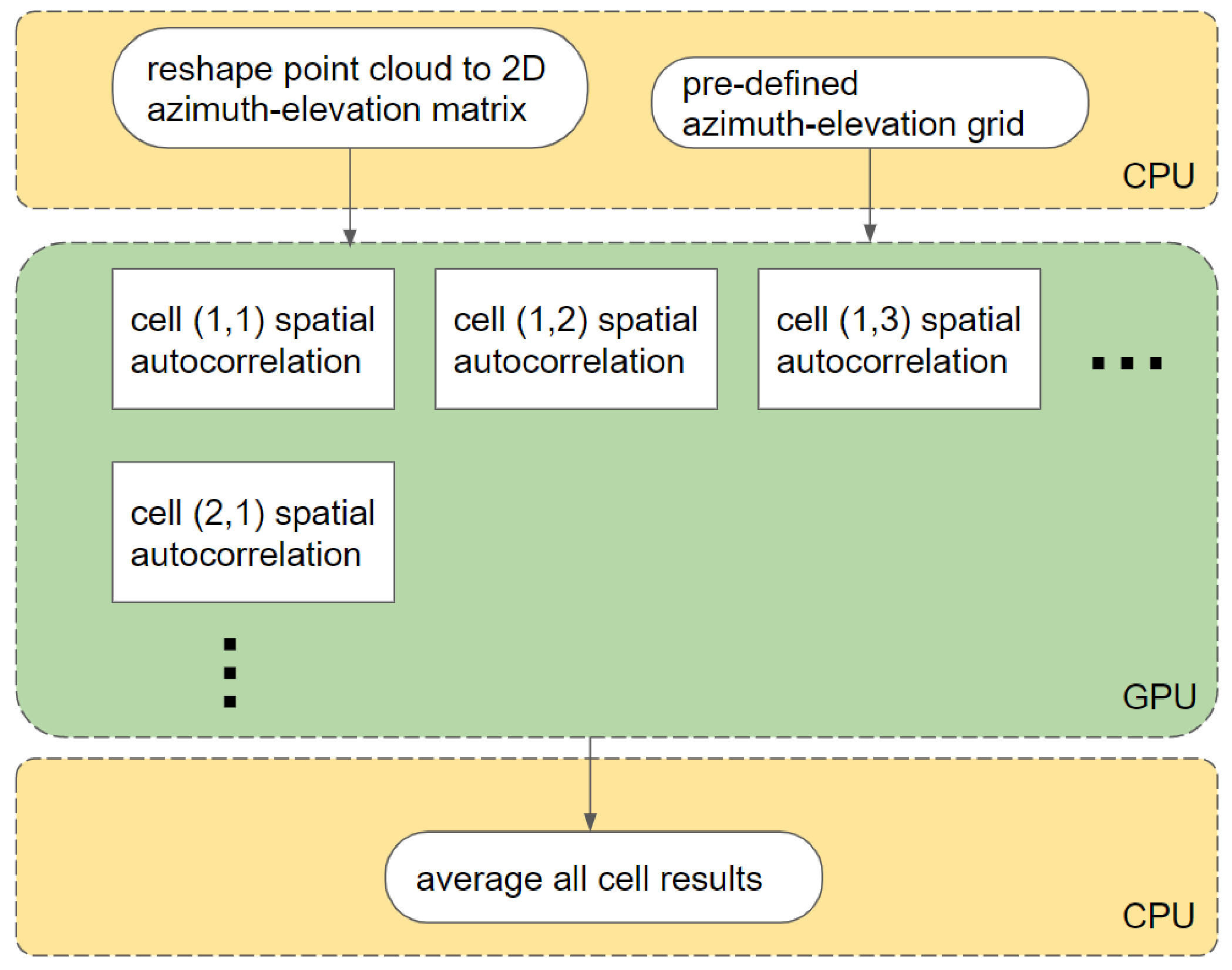}
    \caption{Computation of Spatial Autocorrelation}
    \label{fig:gpu-arch}
\end{figure}

Figure~\ref{fig:gpu-arch} demonstrates the GPU-accelerated parallel computation structure.
For each LiDAR data frame, the pointcloud is first reorganized as an $m\times n$ 2-D array before sending to the GPU, where $m$ and $n$ are pre-defined parameters based on the LiDAR's FOV and resolution.
Note that a LiDAR frame does not necessarily have detection at all entries, and the entries without valid detection are set to have a range of 0 which will be excluded from the spatial autocorrelation calculation.
Given the size of the grid $V$ and $H$ as previously defined, the GPU launches $V\times H$ threads in parallel, and each thread computes the weighted spatial autocorrelation of the LiDAR points within the corresponding grid cell. After all threads finish the computation, the results are sent back to the CPU for the final calculation.

\section{Results}
\label{sec:result}
We collect test data with two different LiDAR models which have different spepcifications in almost all aspects from the scanning mechanism to the laser spectrum.
Table~\ref{tab:lidar-param} lists some of the key parameters of the two LiDAR models.
Both LiDARs calculate the distance measurement on a time-of-flight (TOF) basis.

\begin{table}[htb]
\begin{center}
    \caption{Parameters of Test LiDARs}
    \begin{tabular}{ |c|c|c|c| } 
    \hline
    \hspace{1mm} & \textbf{Wavelength} & \textbf{FOV (H$\times$V)} & \textbf{Mounting Orientation} \\
    \hline
    LiDAR 1  & 905nm & $360^\circ\times 40^\circ$ & Surrounding \\
    \hline
    LiDAR 2  & 1550nm & $120^\circ\times 25^\circ$ & Forward-looking \\
    \hline
    \end{tabular}
    \label{tab:lidar-param}
\end{center}
\end{table}

Several Navistar International LT625 trucks equipped with both LiDAR models is used for data collection on public road.
All LiDARs are mounted in an exposed manner, i.e., no windshield or other secondary fascia in front of the LiDAR. 
Each truck is also equipped with multiple cameras oriented to various directions. The cameras are synchronized with the LiDARs and the camera images are recorded in addition to the LiDAR data as reference.
We have accumulated a total of over 230 unit-hours and 10,000 unit-miles of road data with a combination of conditions covering different aspects, including various time of day such as daytime, nighttime, dusk and dawn, various weather conditions such as clear day, rainy and foggy, and various surroundings such as highway, local road, test track and parking lot.
Both LiDARs output pointcloud at a 10Hz rate, leading to a total amount of over 828k frames of pointcloud data. We calculate the pointcloud quality metric once every second, i.e., once every 10 frames of data. Since the scenarios that produces noise or anomalous LiDAR pointcloud, such as rains and fogs, can typically last for some time in a continuous manner, we are still able to capture the anomalous LiDAR pointcloud without losing much information while reducing the effort to go through the test dataset. 
A summary of the dataset is given in Table~\ref{tab:dataset}.

\begin{table}[h!]
    \begin{center}
        \caption{Portfolio of Test Data (\% of Hours)}
        \begin{tabular}{ |c|c|c|c|c|c| } 
        \hline
        \multicolumn{2}{|c|}{\textbf{Time of Day}} & \multicolumn{2}{|c|}{\textbf{Weather Conditions}} & \multicolumn{2}{|c|}{\textbf{Surroundings}} \\\hline
        Daytime      & 88.25\%          & Clear       & 93.10\%         & Highway                      &       73.40\% \\\hline
        Nighttime    & 11.52\%          & Rainy       & 6.07\%         & Local Road                   &       8.51\% \\\hline
        Dusk/Dawn    & 0.23\%          & Foggy       & 0.83\%         & Track/Lot    &       18.09\% \\\hline
        \end{tabular}
        \label{tab:dataset}
    \end{center}
\end{table}

For the rest of this section, we select a few typical scenarios and analyze in detail to showcase the performance of the proposed method, as well as providing an overview of the method's performance over the entire test data set.

\subsection{Scenario I: Electromagnetic Interference (EMI)}
\begin{figure}[htb]
    \centering
    \includegraphics[width=0.25\textwidth]{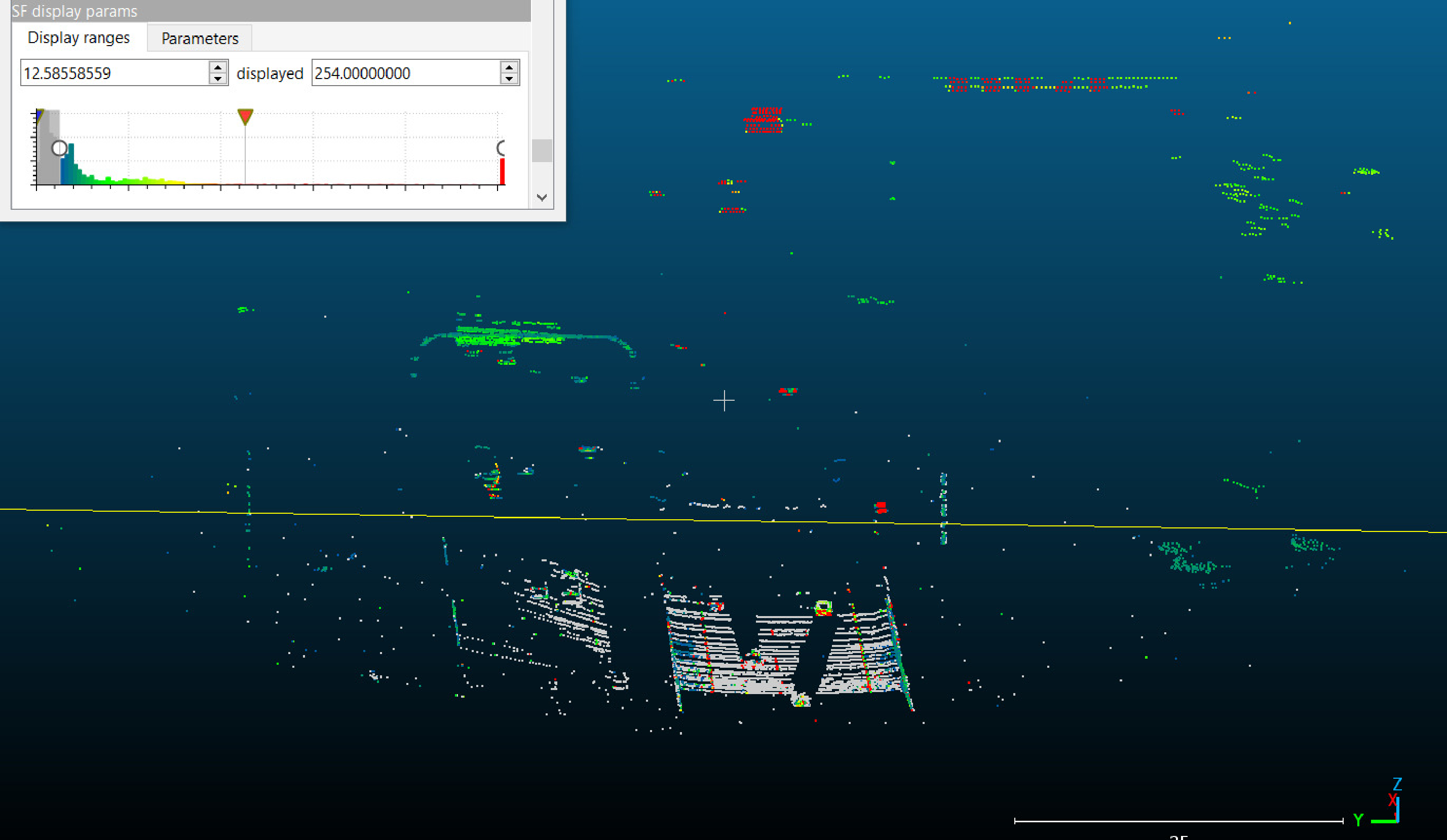}
    \caption{LiDAR Pointcloud with EMI}
    \label{fig:emi-pc}
\end{figure}

In this scenario, one unit of LiDAR 2 with defected EMI shielding passes through a cellular signal tower, generating a large amount of low-intensity noise points. As shown in Figure~\ref{fig:emi-pc}, the noise points are randomly and sparsely distributed over the LiDAR FOV and in general have low intensity values (marked gray together with some points from the road surface).

\begin{figure}[htb]
    \centering
    \subfigure[metric score]{
    \includegraphics[width=0.22\textwidth]{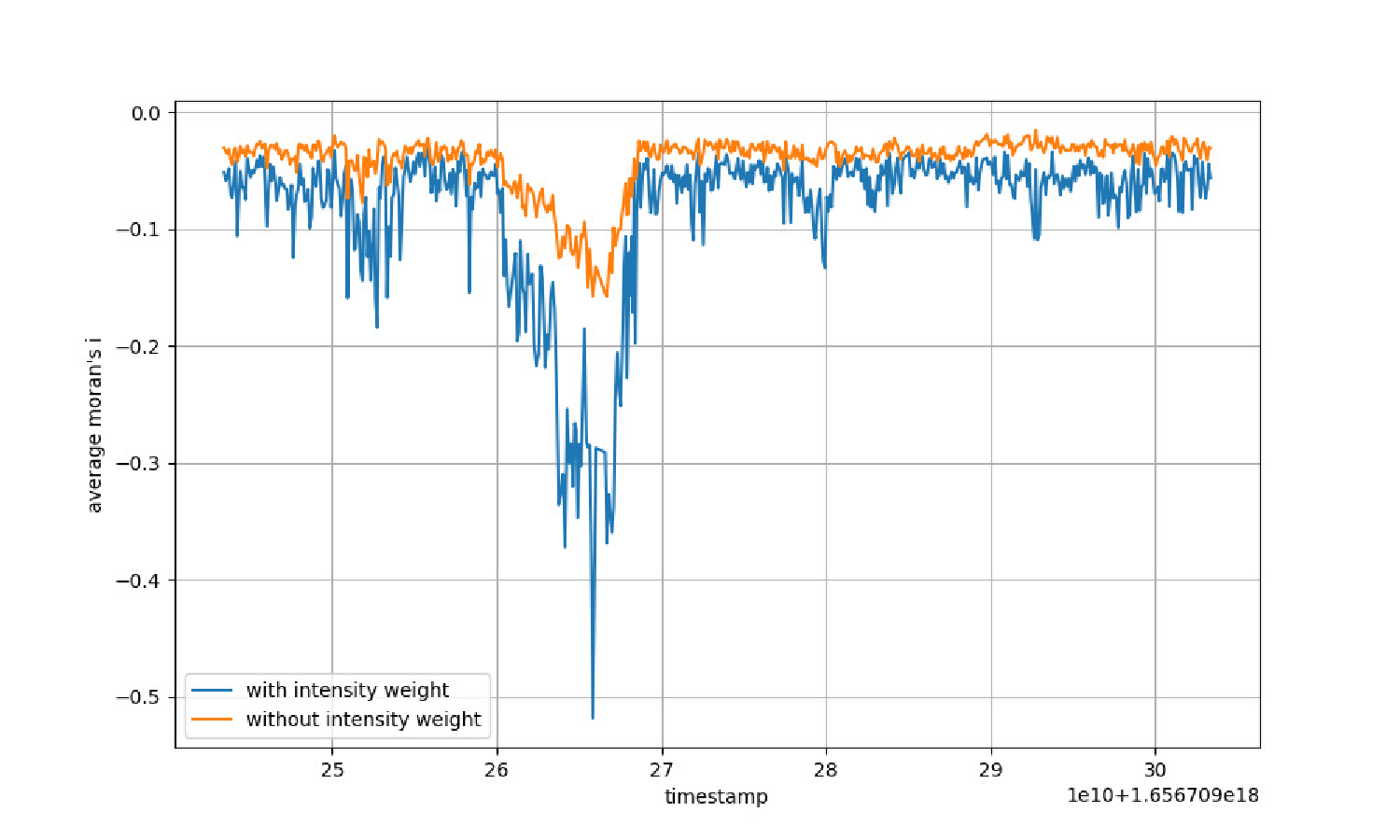}
    \label{fig:emi-autocorr}
    }
    \subfigure[range variance]{
    \includegraphics[width=0.22\textwidth]{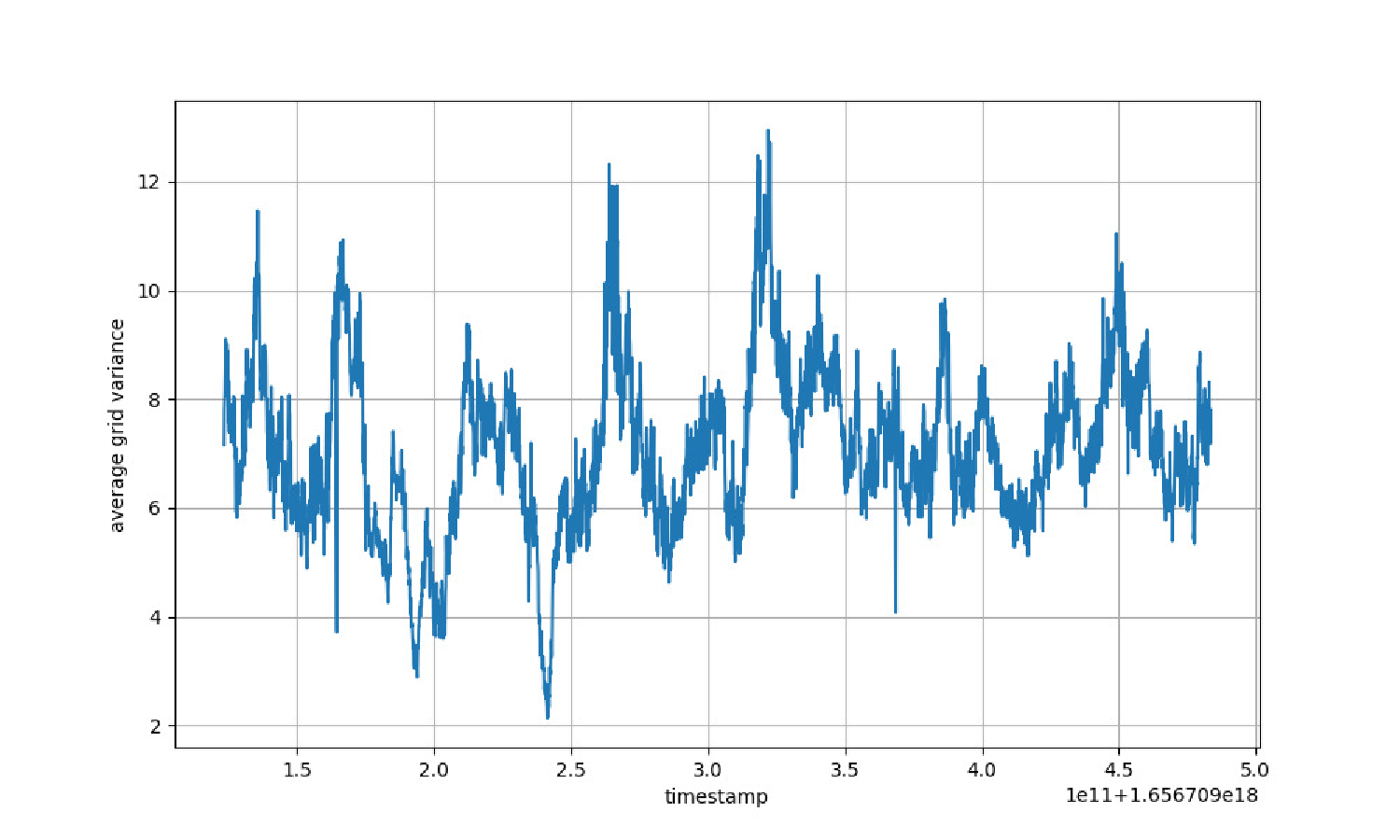}
    \label{fig:emi-stdev}
    }
    \caption{Pointcloud Quality Score of Scenario I Over Time}
\end{figure}

Figure~\ref{fig:emi-autocorr} shows the proposed LiDAR pointcloud quality metric over time. 
To demonstrate how the spatial autocorrelation and the intensity weight multiplier contribute to the overall metric respectively, the orange curve shows the spatial autocorrelation over time without the multipilication of the intensity weight, and the blue curve shows the overall quality metric score. Both curves show significant drops for about 10s which corresponds to the duration of the EMI effect. 
In this scenario, the spatial autocorrelation can clearly capture the false positives, and the intensity weight magnifies the gap between the normal and low-quality data frames since the noise points are mostly low-intensity. 
It should be noted that even for normal data frames, the intensity weight scales the spatial autocorrelation since there are always points with intensity values below the reference intensity. 

As comparison, Figure~\ref{fig:emi-stdev} shows the averaged range variance of all the grid cells over the same data segment.
While the EMI affected pointcloud does lead to a peak in the range variance, there are other peaks when the pointcloud is normal, and the peak at the EMI effect is not significant enough to distinguish the pointcloud frame.
Therefore, the range variance is not a suitable detector for anomalous pointcloud frames.

\begin{figure}[htb]
    \centering
    \subfigure[without EMI]{
    \includegraphics[width=0.22\textwidth]{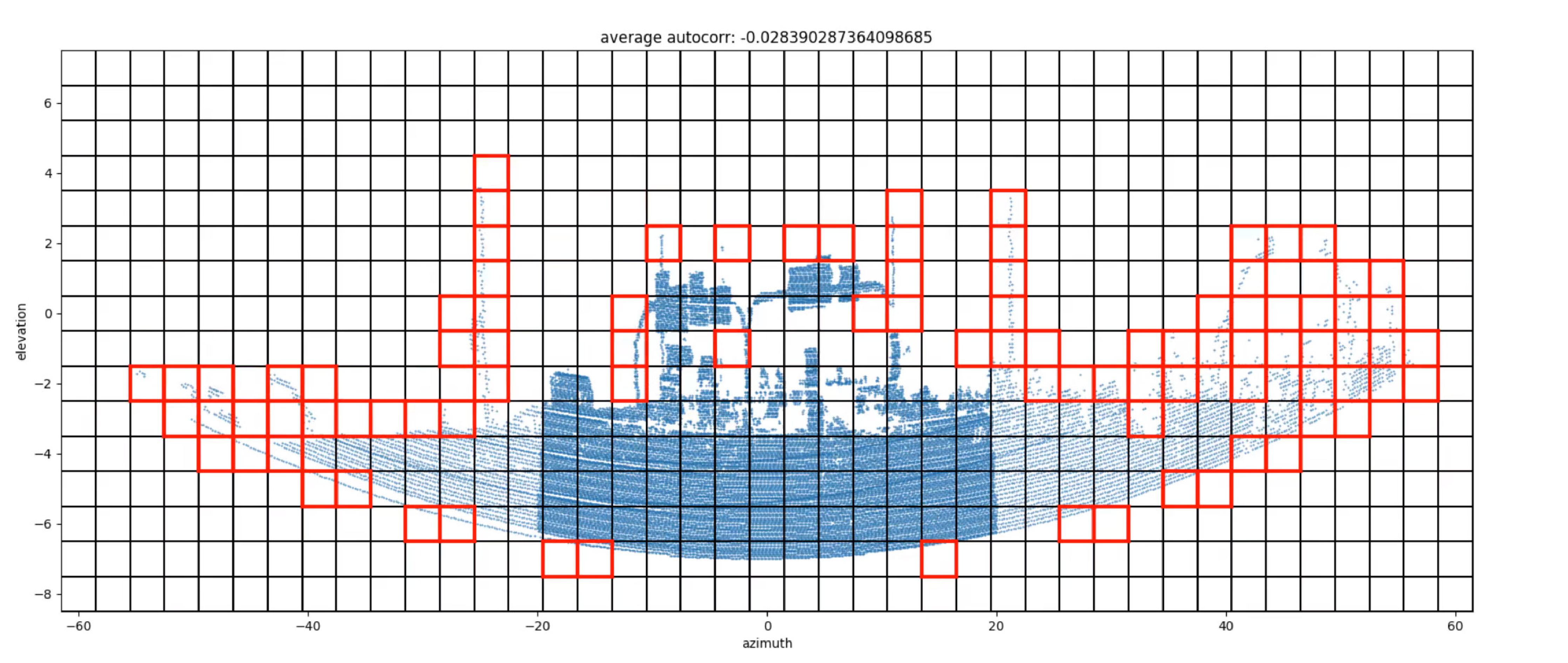}
    \label{fig:grid-without-emi}
    }
    \subfigure[with EMI]{
    \includegraphics[width=0.22\textwidth]{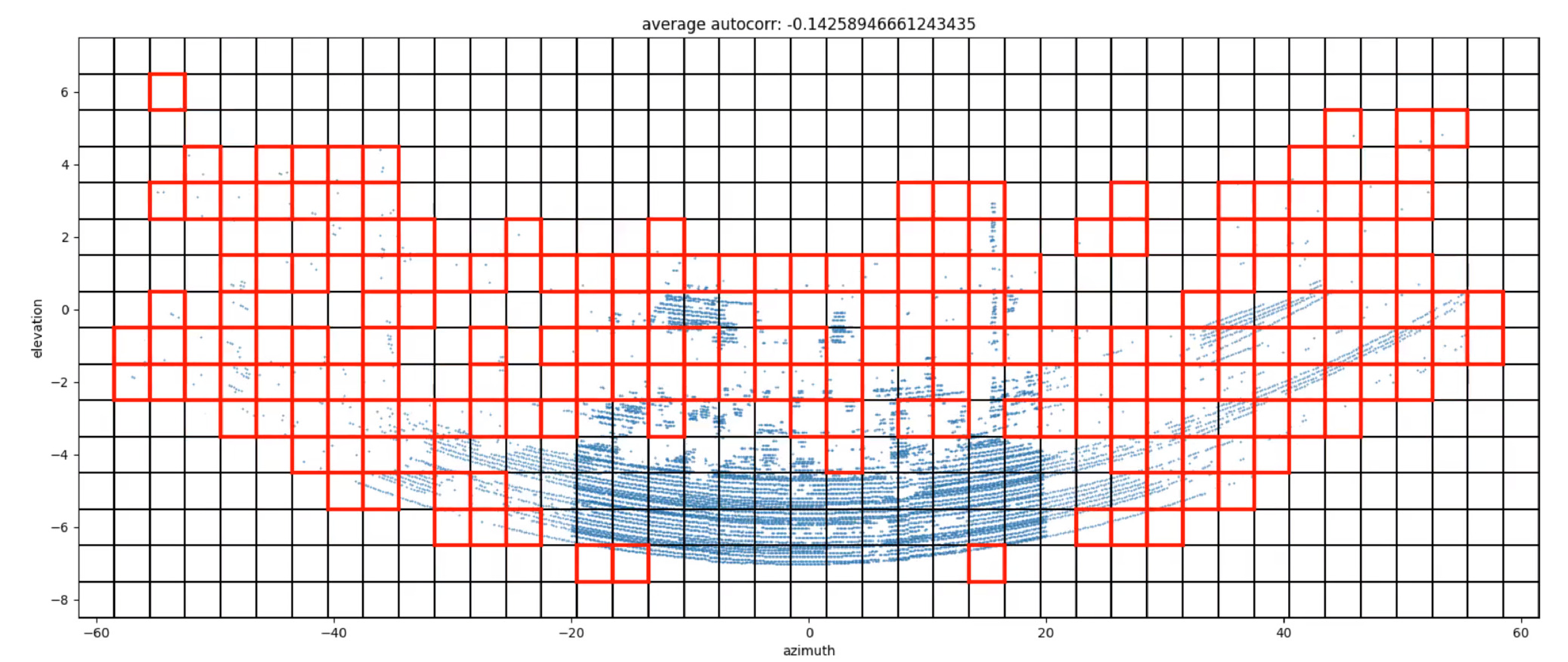}
    \label{fig:grid-with-emi}
    }
    \caption{Two Frames of LiDAR Image Grid during Scenario I}
    \label{fig:emi-grid}
\end{figure}

In Figure~\ref{fig:emi-grid} we showcase two frames of the LiDAR image grid during the scenario, where Figure~\ref{fig:grid-without-emi} captures an instance without any EMI effect and Figure~\ref{fig:grid-with-emi} is one exemplary image grid when the EMI effect is in place, respectively.
The grid cells with red edges have low unweighted spatial autocorrelation values.
As can be seen in Figure~\ref{fig:grid-without-emi}, individual grid cells may have low spatial autocorrelation values occasionally especially at the edge of FOV or when objects with small reflection surfaces, such as poles and vegetation, appear in the pointcloud. However, they do not lead to a low total quality metric score since the amount of such type of grid cell is generally small.
On the other hand, anomalies and noise points generates numerous grid cells with low spatial autocorrelation values, as shown in Figure~\ref{fig:grid-with-emi}. As a result, the overall quality metric of the data frame is low.

\subsection{Scenario II: Rain}
\begin{figure}[htb]
    \centering
    \subfigure[LiDAR 1]{
    \includegraphics[width=0.13\textwidth]{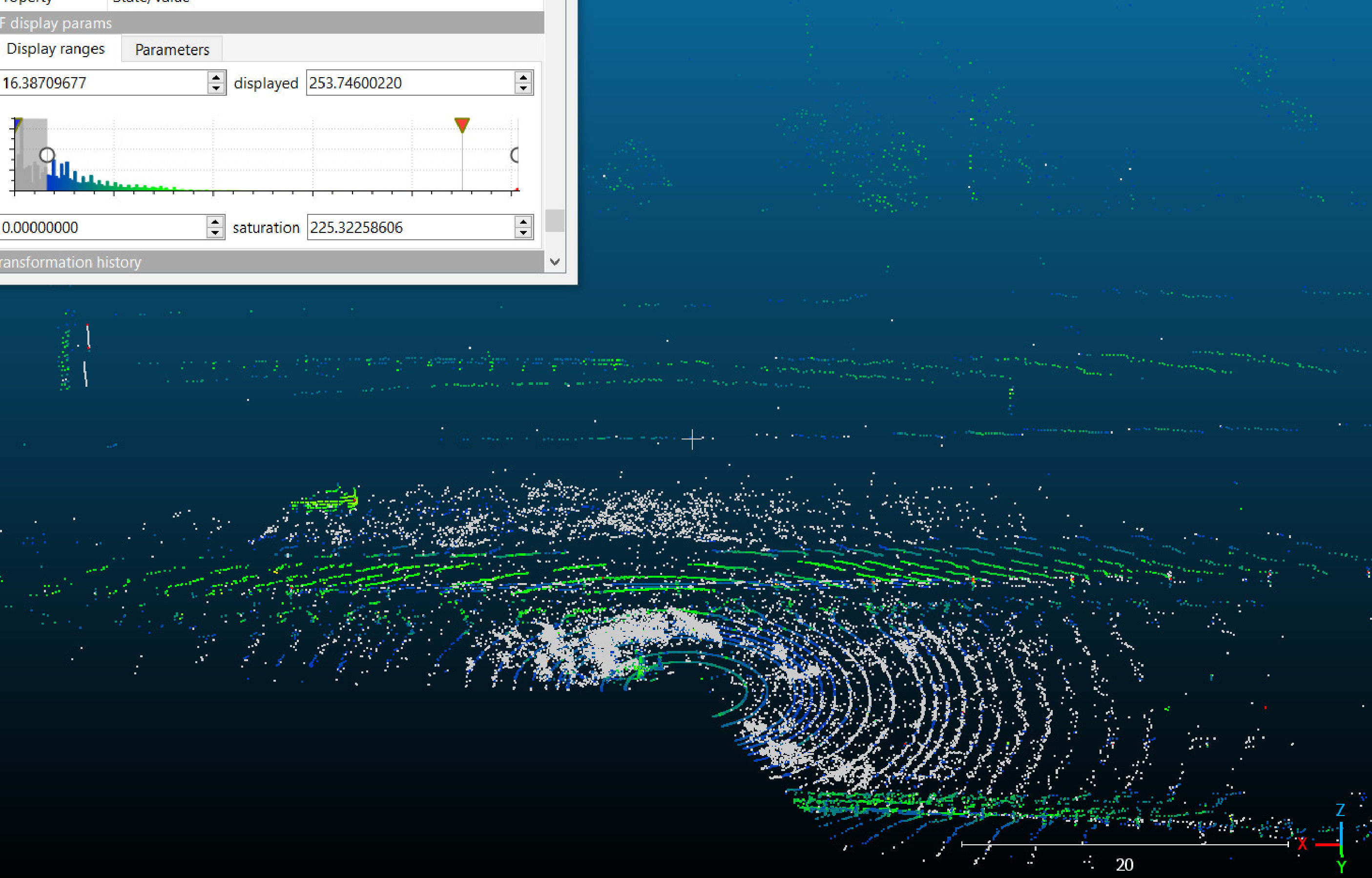}
    \label{fig:pandar-rain-pc}
    }
    \subfigure[LiDAR 2]{
    \includegraphics[width=0.13\textwidth]{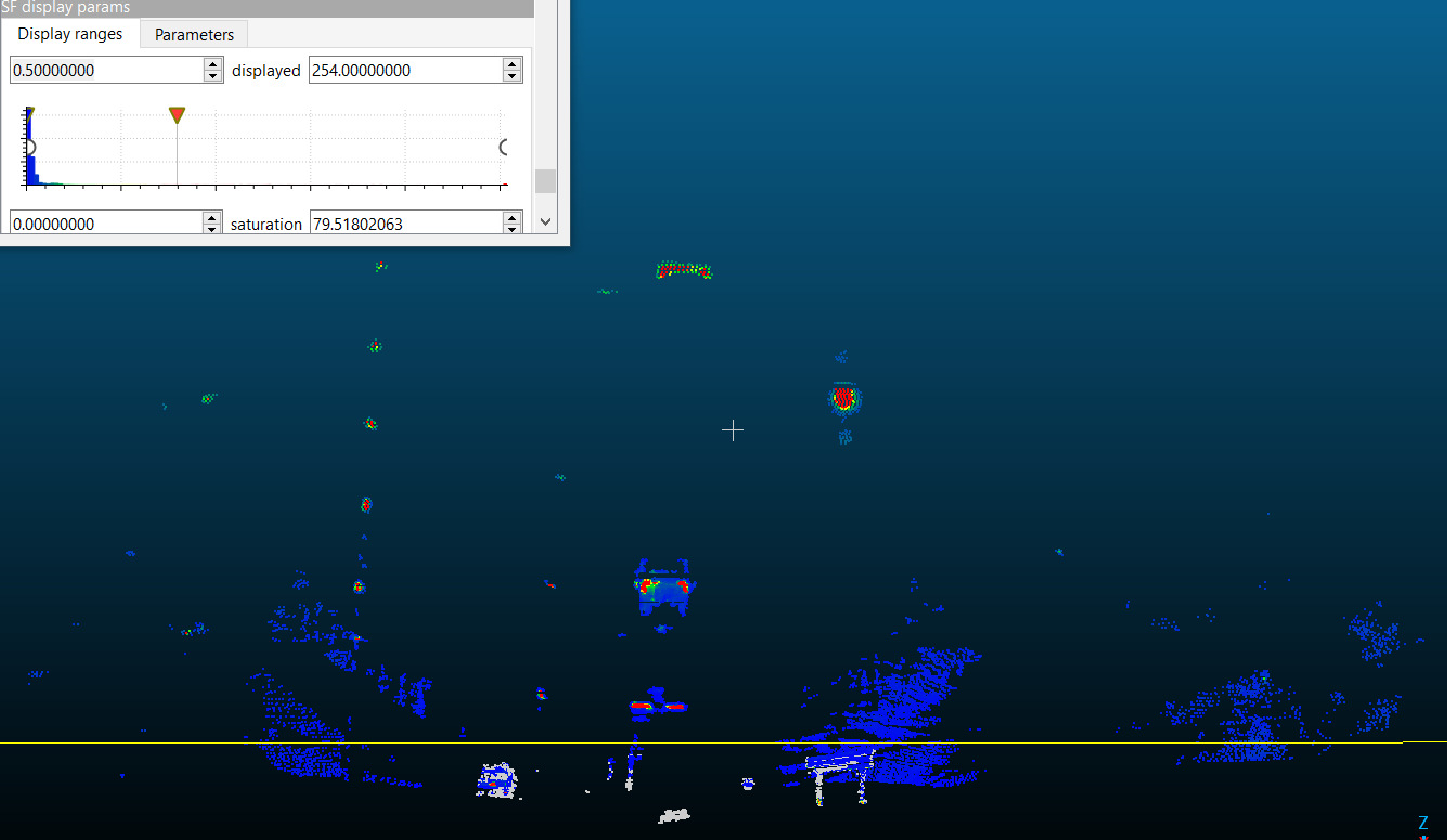}
    \label{fig:falcon-rain-pc}
    }
    \subfigure[reference camera]{
    \includegraphics[width=0.13\textwidth]{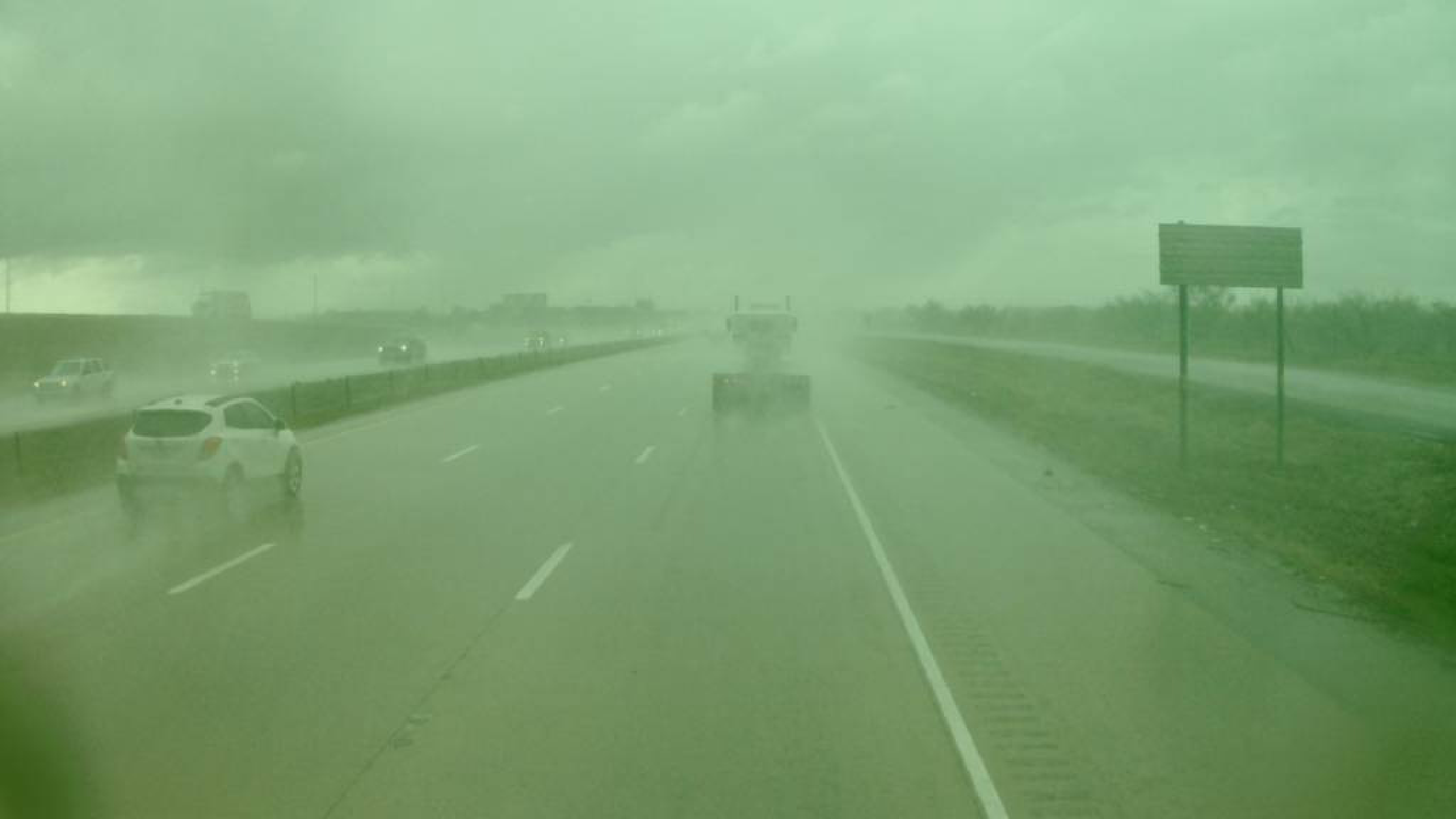}
    \label{fig:rain-cam}
    }
    \caption{Pointcloud and Image in Scenario II}
    \label{fig:pc-rain}
\end{figure}

In this scenario, we investigate a trip segment where both LiDAR models are exposed in heavy rain. 
Figure~\ref{fig:pandar-rain-pc} and Figure~\ref{fig:falcon-rain-pc} show the pointcloud of LiDAR 1 and 2 in the rain scenario, respectively.
A reference camera image is shown in Figure~\ref{fig:rain-cam}.
As demonstrated in the figures, LiDARs with 905nm laser wavelength are more likely to see scattered noise points from rain droplets and water splashes; 
LiDARs with 1550nm laser wavelength generates pointcloud where both the point density and intensity are significantly reduced due to signal absorption, as well as clustered noise points at close ranges.

\begin{figure}[htb]
    \centering
    \subfigure[LiDAR 1]{
    \includegraphics[width=0.22\textwidth]{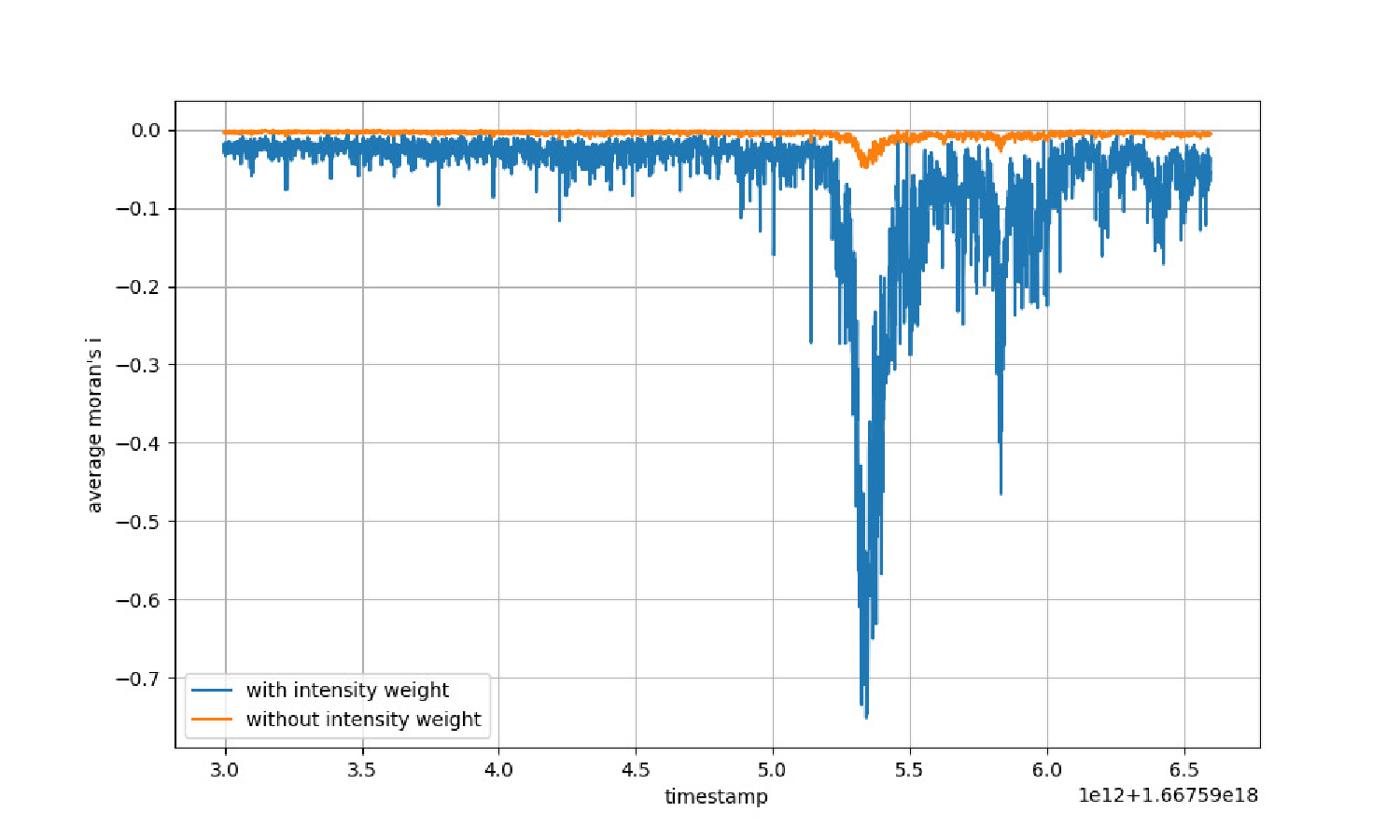}
    \label{fig:pandar-autocorr-rain}
    }
    \subfigure[LiDAR 2]{
    \includegraphics[width=0.22\textwidth]{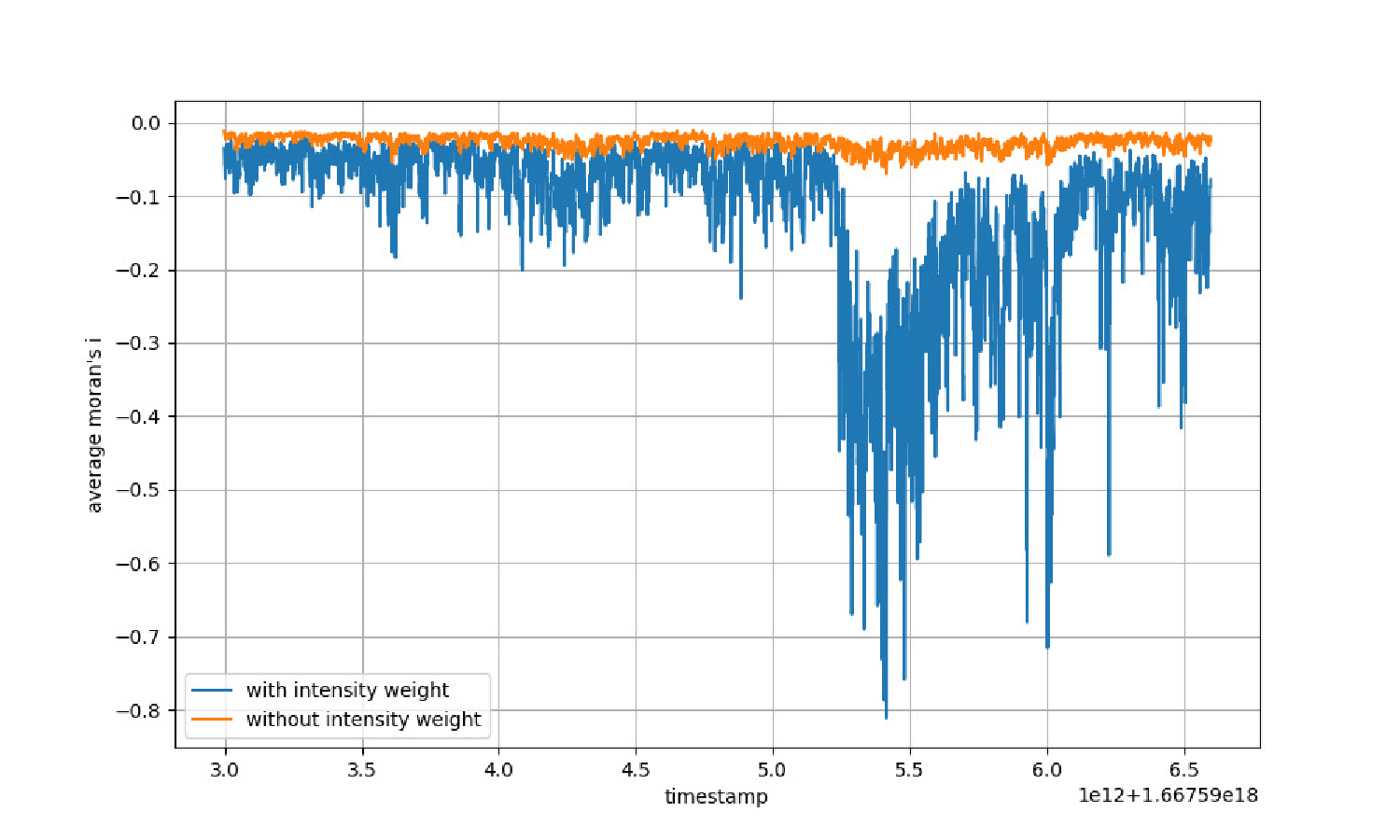}
    \label{fig:falcon-autocorr-rain}
    }
    \caption{Pointcloud Quality Score of Scenario II Over Time}
    \label{fig:autocorr-rain}
\end{figure}

Figure~\ref{fig:pandar-autocorr-rain} gives both the quality metric score and the unweighted spatial autocorrelation from LiDAR 1 during the test. Due to the scattered pattern of the noise seen by the 905nm LiDAR, even the unweighted spatial autocorrelation can distinguish the rain scenario well since the scattered noise tends to generate low spatial autocorrelation scores. And since 905nm LiDARs' laser signal also gets attenuated in rain and leading to lower-than-normal intensities, including the intensity weight multiplier may increase the gap between `normal' pointcloud frames and anomalous pointcloud frames.
Figure~\ref{fig:falcon-autocorr-rain} shows the quality metric score and the unweighted spatial autocorrelation from LiDAR 2. It can be seen that the unweighted spatial autocorrelation in general cannot differentiate the pointcloud frames affected by rain, since the points, including noise points, can be well clustered.
The intensity weight multiplier in this case effectively helps to characterize the rain data.  

\subsection{Test Result Overview}
For the LiDAR data frames with low quality metric score outputs, we define a true positive result when there are notable noise/anomalous points in the pointcloud, and a false positive result when no notable noise/anomalous points are found.
In this section, we pick the true and false positive cases by finding pointcloud frames whose quality metric score is less than -0.4.
It should be noted that the quality metric score threshold is merely a bar to filter out the frames of interest from the large amount of test data, and is not meant to be a threshold for real application. 

\begin{figure}[htb]
    \centering
    \subfigure[true positive causes]{
    \includegraphics[width=0.22\textwidth]{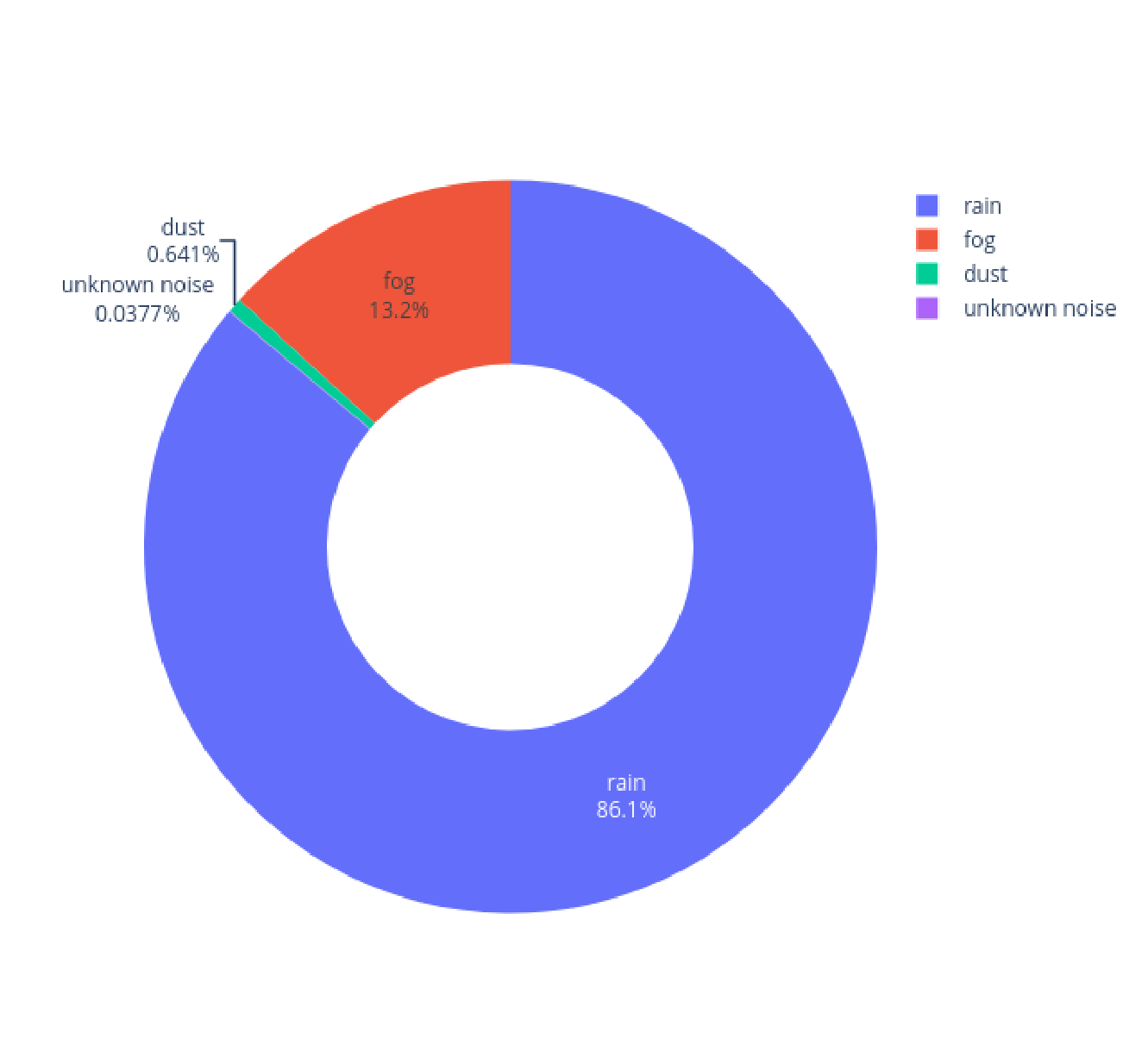}
    \label{fig:tp-distribution}
    }
    \subfigure[false positive causes]{
    \includegraphics[width=0.22\textwidth]{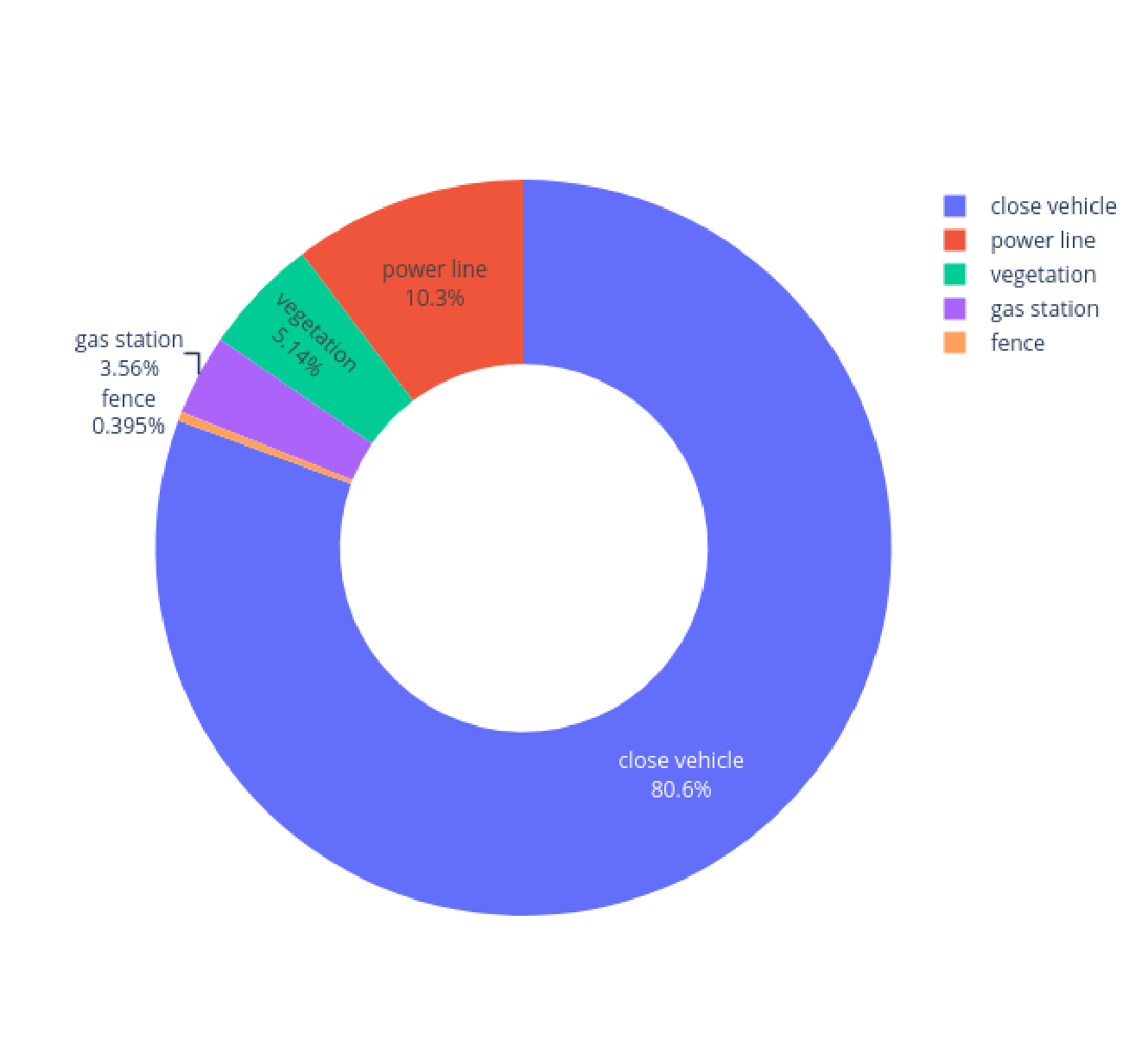}
    \label{fig:fp-distribution}
    }
    \caption{Distribution of Low Score Causes}
    \label{fig:cause-distribution}
\end{figure}

Figure~\ref{fig:tp-distribution} and~\ref{fig:fp-distribution} show the breakdown of causes that lead to true positive and false positive results, respectively.
Among over 82.8k frames of LiDAR pointcloud checked, the amount of frames labeled as true positive is about 16k, which are mainly caused by rain, fog and dust. There are some true positive cases where the noise/anomaly source cannot be identified from the reference camera (unknown noise), which we believe are likely caused by sunlight interference or other reasons.
On the other hand, there are about 250 frames in the test dataset labeled as false positive. 
Typical objects/scenarios that result in false positive pointcloud frames include close vehicles passing by the ego vehicle in adjacent lanes, power lines, and vegetation, which contribute over 95\% of the false positive cases.

\begin{figure}[htb]
    \centering
    \subfigure[close vehicle]{
    \includegraphics[width=0.12\textwidth]{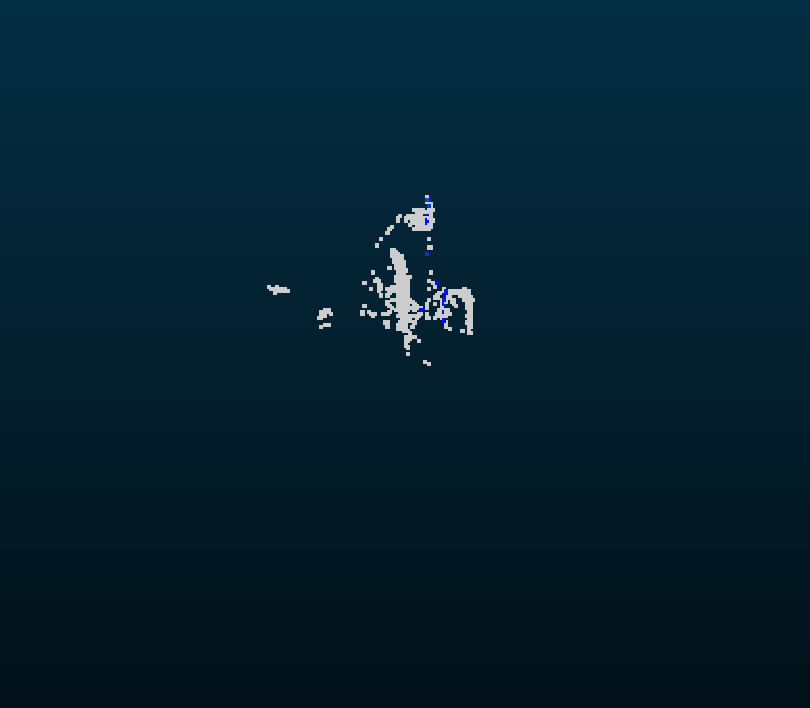}
    \label{fig:close-vehicle-pc}
    }
    \subfigure[power line]{
    \includegraphics[width=0.12\textwidth]{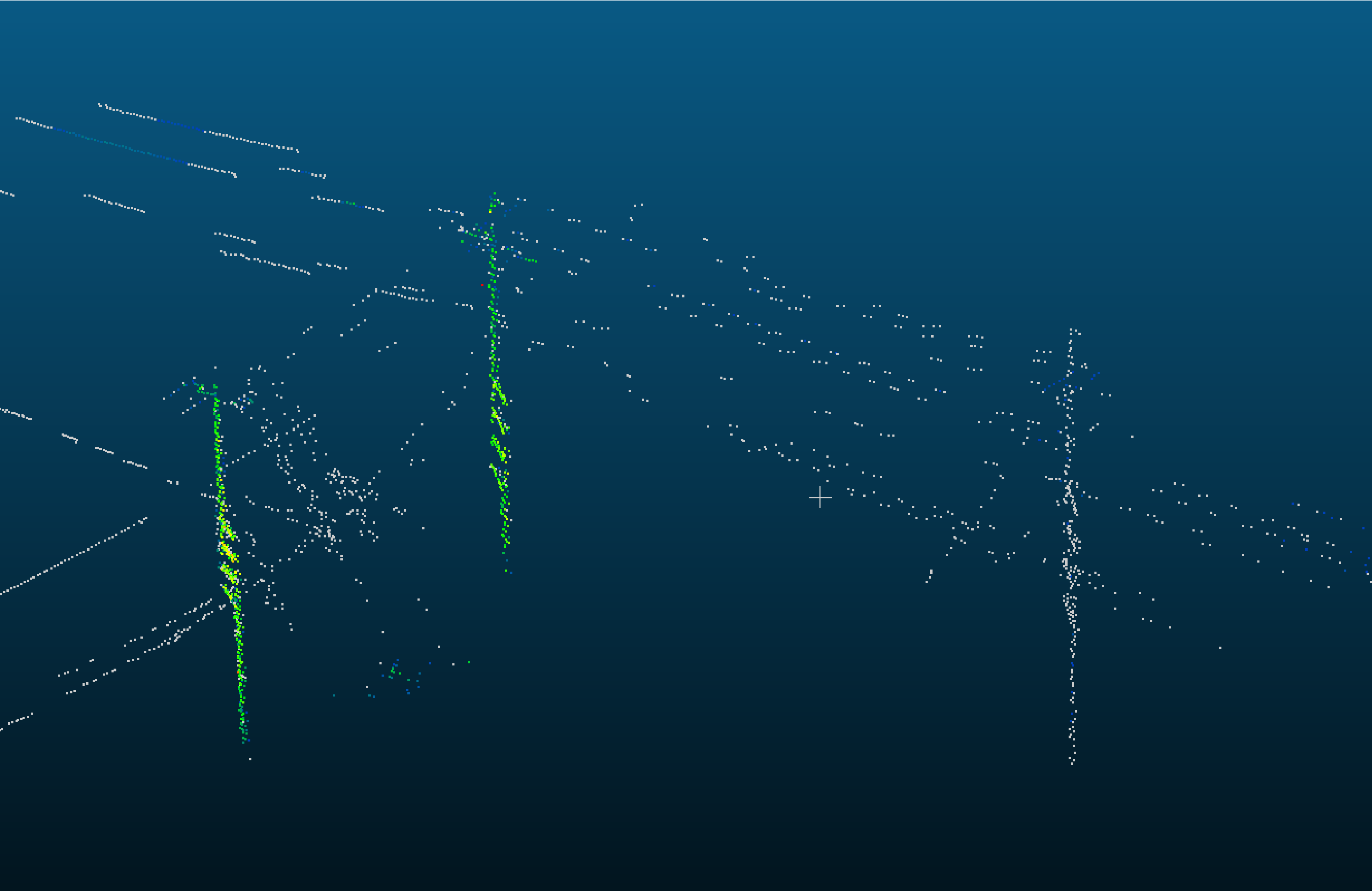}
    \label{fig:power-line-pc}
    }
    \subfigure[vegetation]{
    \includegraphics[width=0.12\textwidth]{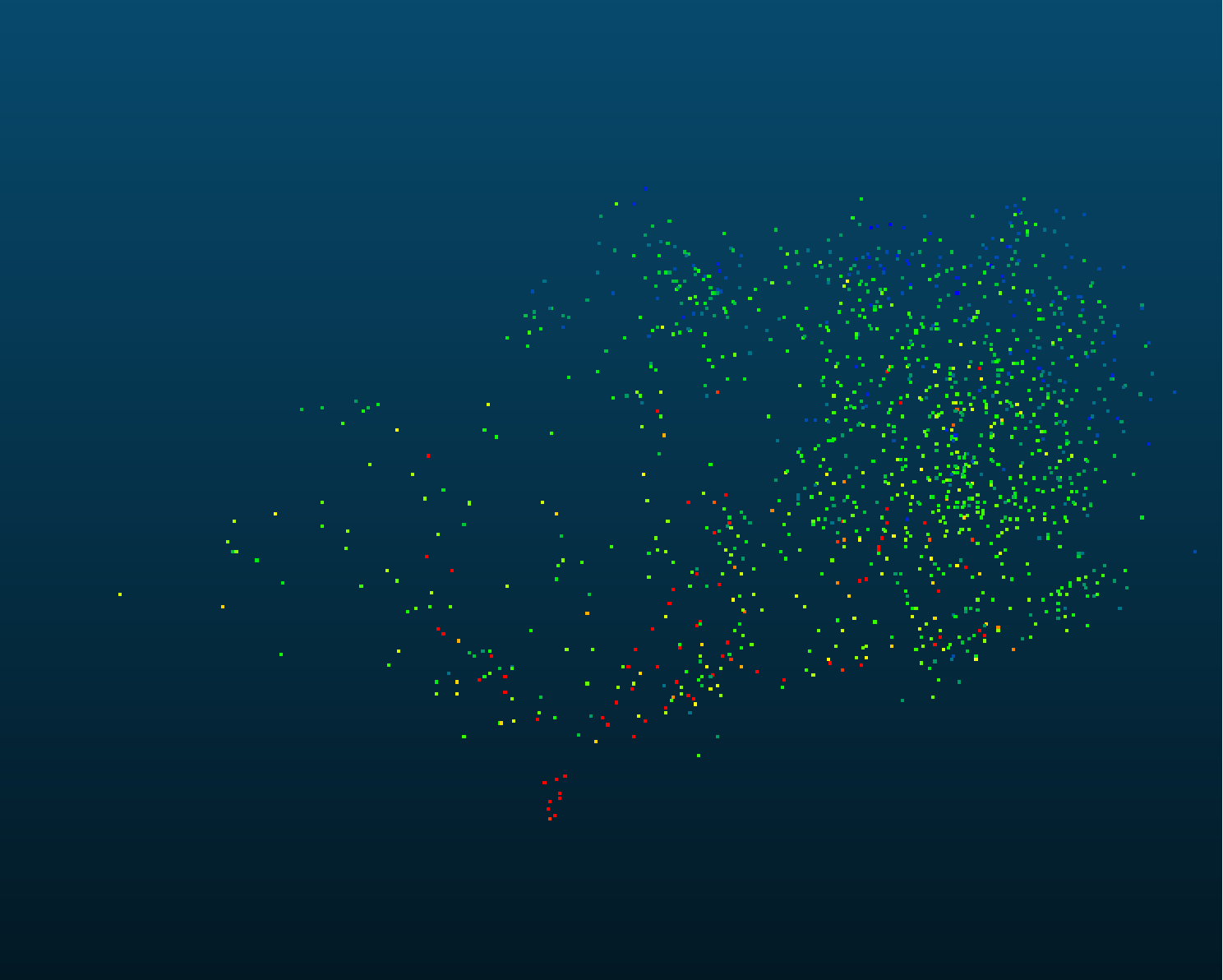}
    \label{fig:tree-pc}
    }
    \caption{Pointcloud of Objects Causing False Positives}
    \label{fig:pc-fp}
\end{figure}

Figure~\ref{fig:pc-fp} provides examples of pointcloud of close vehicles, power lines and vegetation. 
Close vehicles are typically only partially detected by the LiDAR due to the LiDAR's limited FOV. The pointcloud from the partially detected vehicle can be random and scarce, depending on which and how much portion of the vehicle is within the FOV. 
In addition, the LiDAR ranging precision of close vehicles is often degraded due to the multi-reflection between the target and ego vehicle bodies, leaving scattered points over the space. 
The point intensities from close vehicles can also get low due to lasers hitting the smooth vehicle surfaces at large angles of incidence.
All the reasons above make the points from close vehicles similar to noise/anomalous points based on our quality metric definition. 
The power lines and vegetation have small area of reflectance and therefore is in general only partially detected with low signal power reflected to the LiDAR's laser detector. Therefore, points from power lines are sparsely distributed as well as having low intensity values, which are close to the characteristics of noise and anomalous points. 

\begin{figure}[htb]
    \centering
    \includegraphics[width=0.22\textwidth]{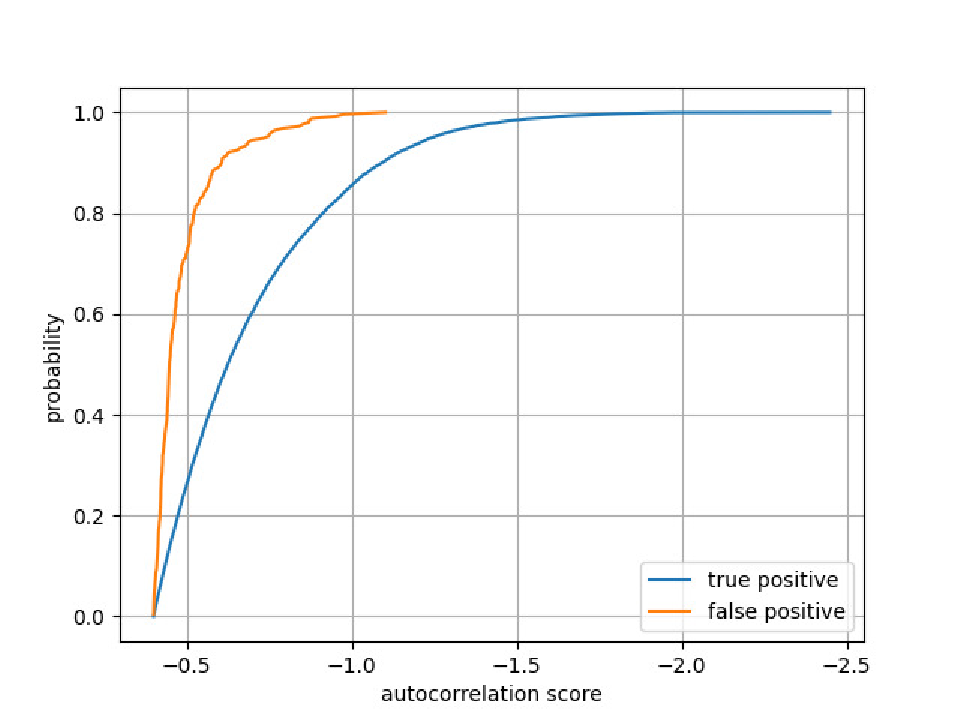}
    \caption{Cumulative Distribution of True/False Positive Case Scores}
    \label{fig:score-cdf}
\end{figure}

Figure~\ref{fig:score-cdf} shows the cumulative distribution of the quality metric scores of all true positive and false positive data frames. 
While there are overlaps between the scores of all true positive and those of all false positive data frames, it is clear that the true positive cases in general have lower scores than the false positive cases.
In our test dataset, over 75\% of the false positive cases have a quality metric score higher than -0.5, while the percentage of the true positive cases that have a score higher than -0.5 is about 25\%. 
Furthermore, while the scores of the true positive cases are distributed in a wide range, it should be noted that the value of the score is related to the severity of the noise/anomaly caused by the source of noise such as rain, fog, dust, etc. 

\begin{figure}[htb]
    \centering
    \subfigure[rain, score=-0.4]{
    \includegraphics[width=0.12\textwidth]{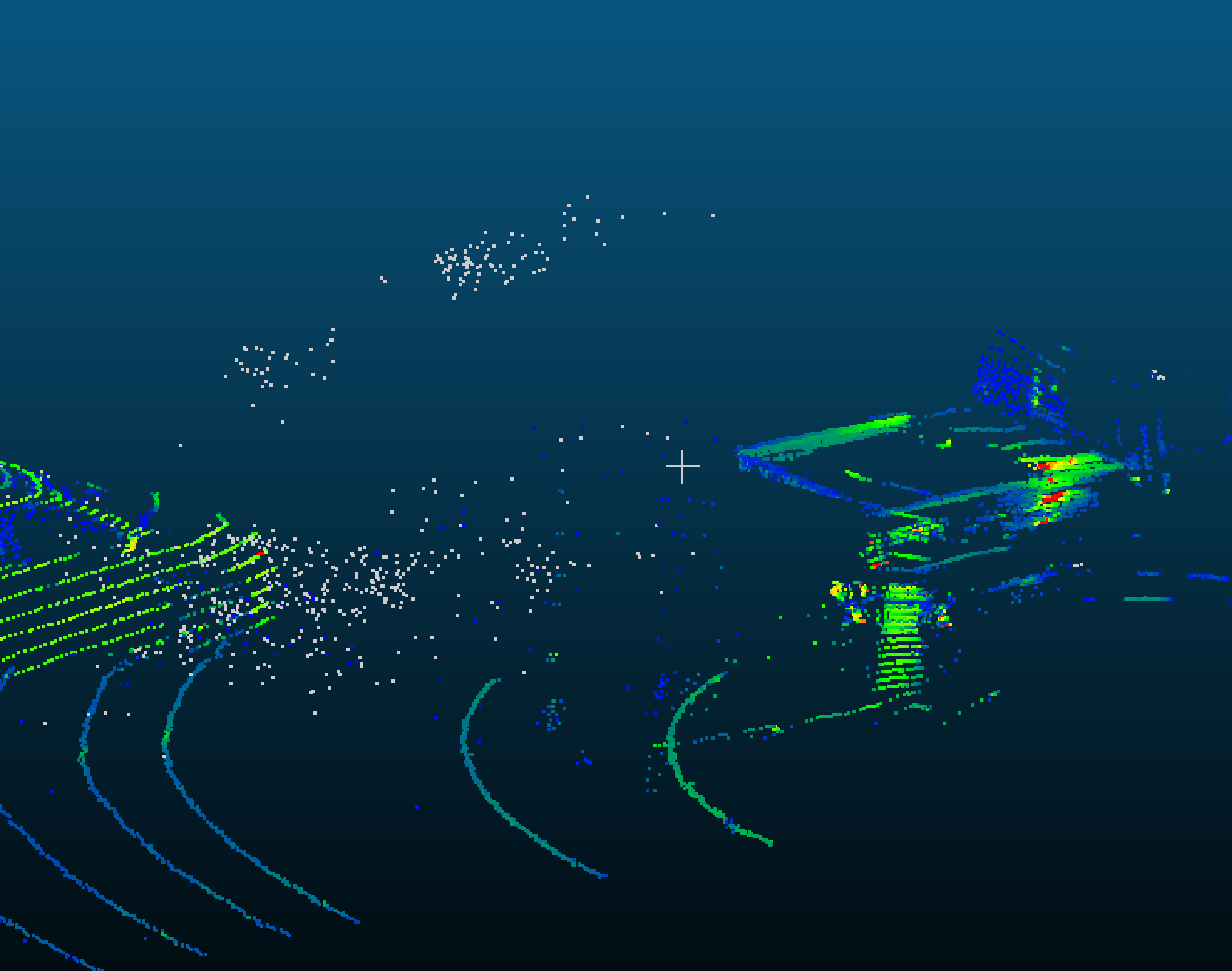}
    \label{fig:rain04}
    }
    \subfigure[rain, score=-0.8]{
    \includegraphics[width=0.12\textwidth]{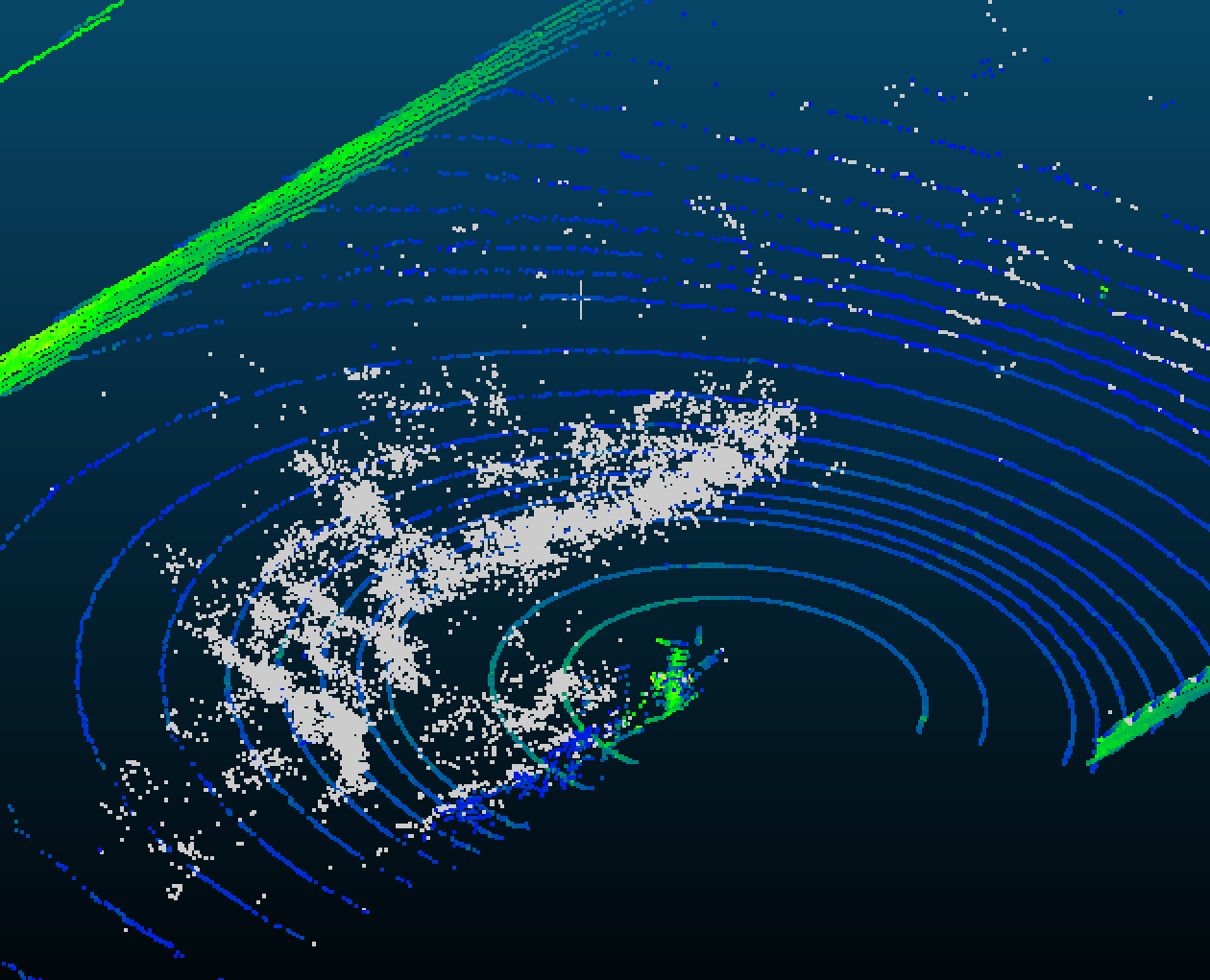}
    \label{fig:rain08}
    }
    \subfigure[rain, score=-1.4]{
    \includegraphics[width=0.12\textwidth]{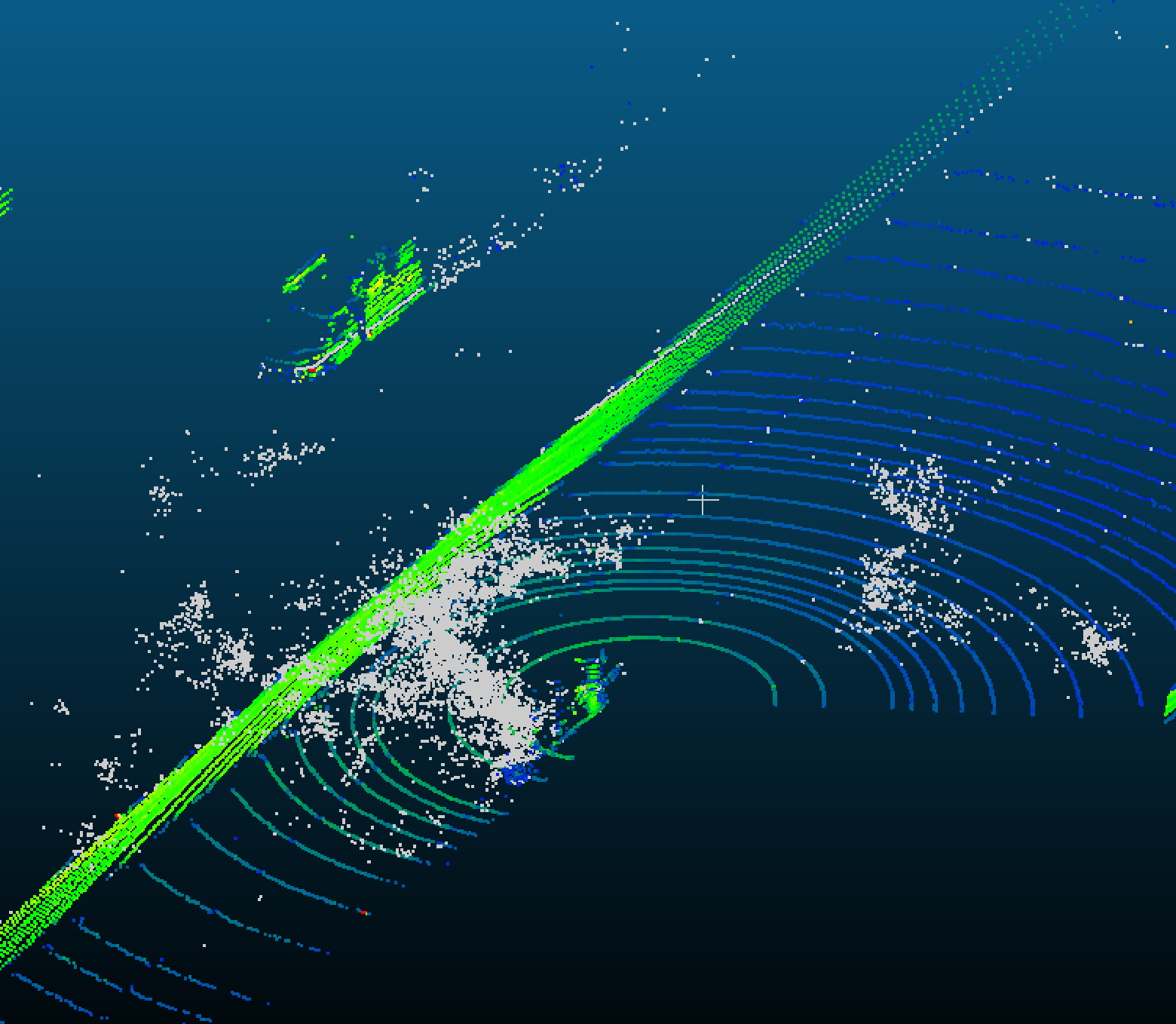}
    \label{fig:rain14}
    }
    \subfigure[fog, score=-0.4]{
    \includegraphics[width=0.12\textwidth]{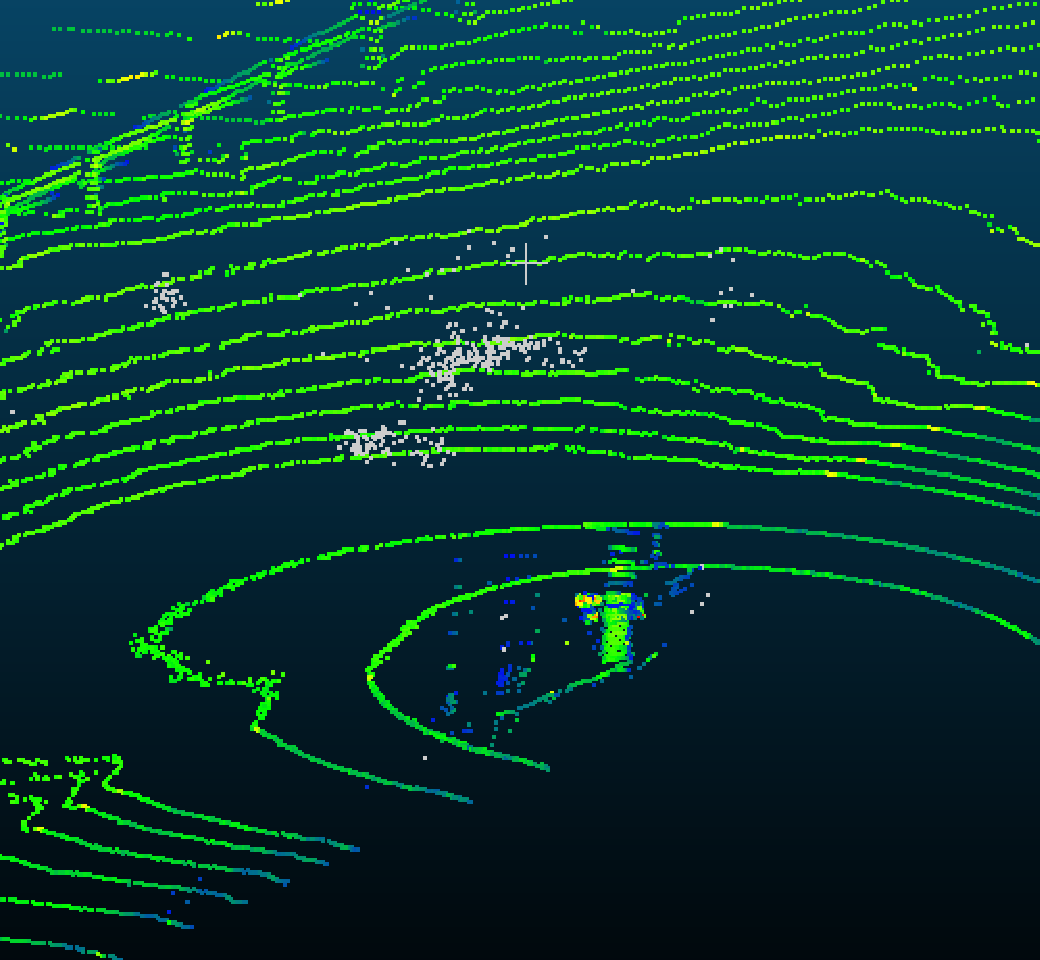}
    \label{fig:fog04}
    }
    \subfigure[fog, score=-0.9]{
    \includegraphics[width=0.12\textwidth]{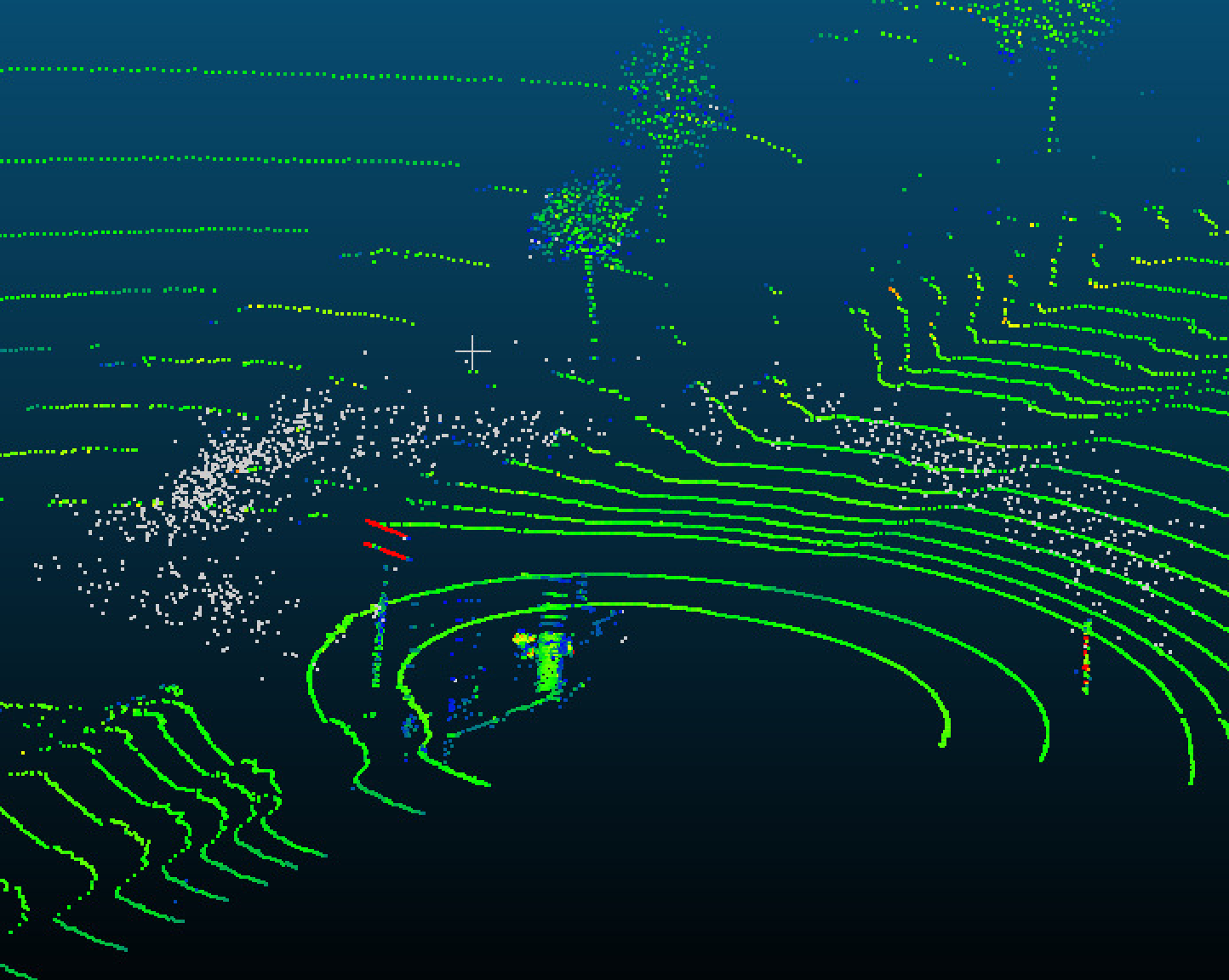}
    \label{fig:fog09}
    }
    \subfigure[fog, score=-2.4]{
    \includegraphics[width=0.12\textwidth]{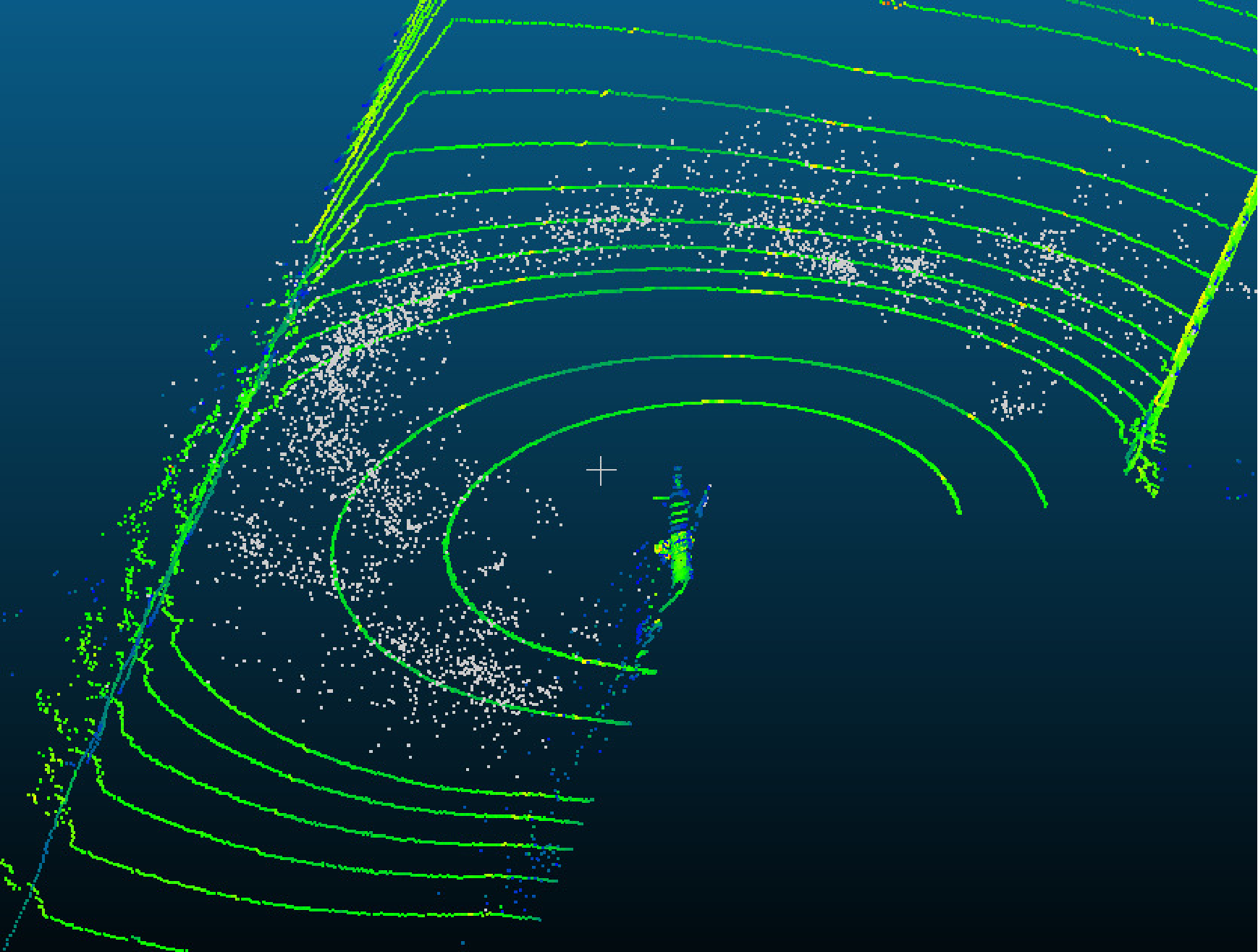}
    \label{fig:fog24}
    }    
    \caption{Pointcloud of True Positive Cases with Different Quality Scores}
    \label{fig:tp-pc-scores}
\end{figure}

Figure~\ref{fig:tp-pc-scores} demonstrates the pointcloud of various rain and fog scenarios and their corresponding quality metric score. The noise points generated from rain and fog are manually marked as gray. 
Sometimes the pointcloud recorded when rain or fog is in presence can have a quality metric score as high as -0.4, however, they typically correspond to light rain or fog scenarios and the amount of noise points is relatively small. 
The pointcloud having a large amount of noise points and a quality metric score going as low as -1.0 or even lower is typically associated with heavy rain or dense fog which may harm the driving safety.
Therefore, for actual application of the proposed method, one may choose a score threshold which best suits their use cases.
For instance, to apply the proposed method to determine whether the vehicle is in an adverse scenario which may be outside the ADS operation domain, one may use a lower detection threshold so that the ADS is not constantly disturbed by false positive detections while the most severe rains and fogs are captured. 
On the other hand, for applications aiming to study the characteristics of noisy and anomalous LiDAR pointcloud, one may choose a high threshold that keeps as many true positive cases as possible and tolerate the increased amount of false positives.

\begin{figure}[htb]
    \centering
    \includegraphics[width=0.22\textwidth]{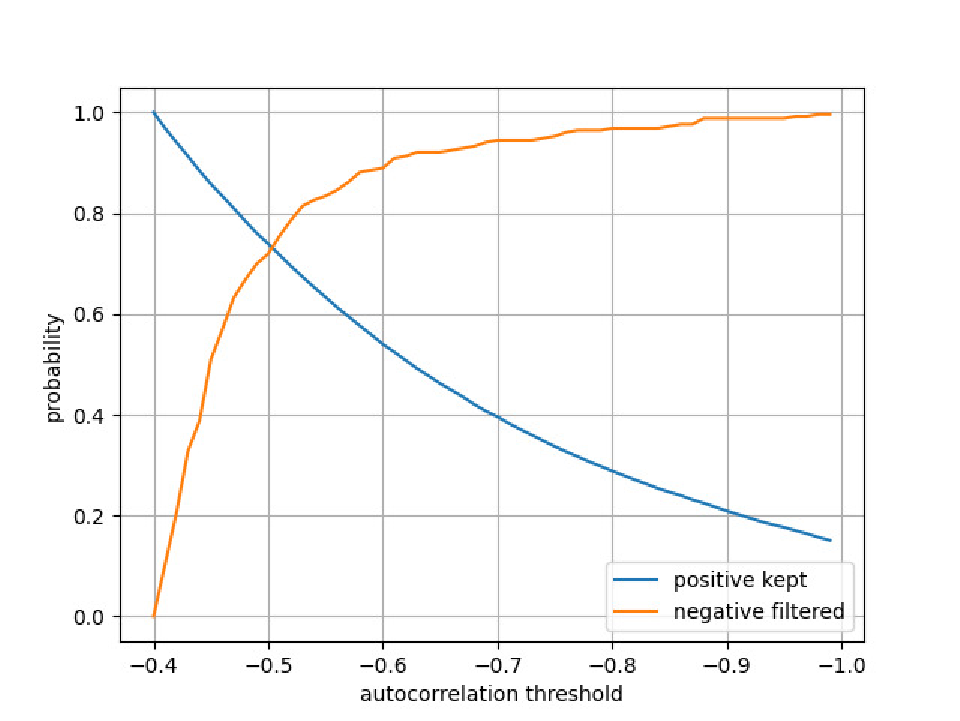}
    \caption{True/False Positive Case Filtering with Different Score Cut-off}
    \label{fig:filter-rate}
\end{figure}

Figure~\ref{fig:filter-rate} uses our test dataset as an example to illustrate the effect of different thresholds. The two curves in the figure show the proportion of true positive cases kept and false positive cases filtered under various score thresholds, respectively. 
With a threshold set to -1.0, one can keep more than 15\% of most severe scenarios while eliminating over 99\% of false positive cases.
On the other hand, a threshold of -0.5 can keep over 70\% of the true positive cases as well as about 25\% of the false positive cases.

\section{Conclusion}
\label{sec:conclusion}
In this paper, we present a novel approach to characterize the noise and anomalies in the LiDAR pointcloud, which is typically caused by adverse environment conditions such as rain, fog, dust, or LiDAR internal component failures. 
To capture the anomalous pointcloud frames, we developed a quality metric score based only on the LiDAR pointcloud characteristics, i.e., the spatial distribution of the points and the intensity values, which does not require any data annotation or training. 
We verified the method with numerous test data collected from public road with various LiDAR physical modalities, and the result proves that the proposed quality metric score can effectively capture the anomalous LiDAR pointcloud caused by different reasons.
There is a wide range of potential applications of the work in this paper, such as monitoring the operation condition of an autonomous driving system in real time, or efficiently selecting the data collected in rain/fog from enormous amount of test data for further analysis.

\bibliographystyle{IEEEtran}
\bibliography{references}

\end{document}